\documentclass[10pt,twocolumn,letterpaper]{article}

\usepackage{style}
\usepackage{times}
\usepackage{epsfig}
\usepackage{graphicx}
\usepackage{amsmath}
\usepackage{amssymb}
\usepackage{xcolor}
\usepackage{subcaption}
\usepackage{array}
\usepackage{multirow}
\usepackage{stfloats}
\usepackage{siunitx}

\newcommand{\dx}{\mathbf{dx}}
\newcommand{\I}{\mathbf{I}}

\usepackage{authblk}

\usepackage[pagebackref=true,breaklinks=true,letterpaper=true,colorlinks,bookmarks=false]{hyperref}

\cvprfinalcopy %

\begin{document}

\title{A Pixel-Based Framework for Data-Driven Clothing\vspace{-10pt}}

\setlength{\affilsep}{0.3em}

\author[1]{Ning Jin}
\author[1]{Yilin Zhu}
\author[1]{Zhenglin Geng}
\author[1,2]{Ronald Fedkiw}
\affil[1]{Stanford University}

\affil[2]{Industrial Light \& Magic}

\makeatletter
\renewcommand\AB@emaillist{\small{\texttt{\{njin19,yilinzhu,zhenglin\}@stanford.edu, fedkiw@cs.stanford.edu}}}
\makeatother

\maketitle
\begin{abstract}
With the aim of creating virtual cloth deformations more similar to real world clothing, we propose a new computational framework that recasts three dimensional cloth deformation as an RGB image in a two dimensional pattern space.
Then a three dimensional animation of cloth is equivalent to a sequence of two dimensional RGB images, which in turn are driven/choreographed via animation parameters such as joint angles.
This allows us to leverage popular CNNs to learn cloth deformations in image space.
The two dimensional cloth pixels are extended into the real world via standard body skinning techniques, after which the RGB values are interpreted as texture offsets and displacement maps.
Notably, we illustrate that our approach does not require accurate unclothed body shapes or robust skinning techniques.
Additionally, we discuss how standard image based techniques such as image partitioning for higher resolution, GANs for merging partitioned image regions back together, etc., can readily be incorporated into our framework.

\end{abstract}

\section{Introduction}\label{sec:intro}
Virtual clothing has already seen widespread adoption in the entertainment industry including feature films (\eg, Yoda~\cite{Bridson:2002}, Dobby~\cite{Bridson:2003}, Monsters, Inc.~\cite{Baraff:2003}), video games (\eg,  \cite{deAguiar:2010,Kavan:2011,Kim:2013,Kim:2012,Muller:2010,Muller:2007}), and VR/AR and other real-time applications (\eg, \cite{Hilsmann2009,1366175,Wang:2010,Xu:2014}). 
However, its potential use in e-commerce for online shopping and virtual try-on is likely to far surpass its use in the entertainment industry especially given that clothing and textiles is a three trillion dollar industry\footnote{\url{https://fashionunited.com/global-fashion-industry-statistics}}.
Whereas games and real-time applications can use lower quality cloth and films have the luxury of a large amount of time and manual efforts to achieve more realistic cloth, successful e-commerce clothing applications demand high quality predictive clothing with fast turnaround, low computational resource usage, and good scalability.

Although there have been many advances in cloth simulation, the ability to match real cloth of a specific material, especially with highly detailed wrinkling, hysteresis, etc. is rather limited. 
Moreover, contact and collision approaches typically lack physical accuracy due to unknown parameters dependent on a multitude of factors even including body hair density and garment thread friction. %
Thus, while embracing simulation and geometric techniques wherever possible, we pursue a new paradigm approaching clothing on humans in a fashion primarily driven by data at every scale.
This is rather timely as 3D cloth capture technology is starting to seem very promising~\cite{Chen:2015,Pons-Moll:2017,robertini2014}. 

Motivated by a number of recent works that view cloth deformations as offsets from the underlying body \cite{DRAPE2012,Neophytou:2014,Pons-Moll:2017,Yang_2018_ECCV} as well as the recent phenomenal impact of convolutional neural networks for image processing \cite{He2015,Alexnet2012,Long_2015_CVPR,faster_rcnn2017,unet2015,vgg2014}, we recast cloth deformation as an image space problem.
That is, we shrink wrap a cloth mesh onto the underlying body shape, viewing the resulting shrink-wrapped vertex locations as pixels containing RGB values that represent displacements of the shrink-wrapped cloth vertices from their pixel locations in texture and normal coordinates.
These cloth pixels are barycentrically embedded into the triangle mesh of the body, and as the body deforms the pixels move along with it; however, they remain at fixed locations in the pattern space of the cloth just like standard pixels on film. 
Thus, cloth animation is equivalent to playing an RGB movie on the film in pattern space, facilitating a straightforward application of CNNs.
Each cloth shape is an image, and the animation parameters for joint angles are the choreography that sequences those images into a movie of deforming cloth. 

Although we leverage body skinning \cite{scape2005,Kavan:2007,Kavan:2005,SMPL:2015,Magnenat-Thalmann:1988} to move the cloth pixels around in world space, we are not constrained by a need to ascertain the unclothed body shape accurately as other authors aim to \cite{Neophytou:2014,Pons-Moll:2017}. %
Of course, an accurate unclothed body shape might reduce variability in the cloth RGB image to some degree, but it is likely that CNN network efficacy will advance faster than the technology required to obtain and subsequently accurately pose unclothed body shapes. 
Even if consumers were willing to provide more accurate unclothed body data or inferences of their unclothed body forms improve, it is still difficult to subsequently pose such bodies to create accurate shapes governed by animation parameters such as joint angles. 
In contrast, we demonstrate that CNNs can learn the desired clothing shapes even when unclothed body shapes are intentionally modified to be incorrect,
thus providing some immunity to problematic skinning artifacts (\eg, candy wrapper twisting~\cite{Jacobson:2011,Kavan:2007,Kavan:2005}).

\section{Related Work}\label{sec:related_works}
\textbf{Skinning:}
Linear blend skinning (LBS) \cite{lander1998,Magnenat-Thalmann:1988} is perhaps the most popular skinning scheme used in animation software and game engines. 
Although fast and computationally inexpensive, LBS suffers from well-known artifacts such as candy wrapper twisting, elbow collapse, etc., and many works have attempted to alleviate these issues, \eg, spherical blend skinning (SBS)~\cite{Kavan:2005}, dual-quaternion skinning (DQS)~\cite{Kavan:2007}, stretchable and twistable bones skinning (STBS)~\cite{Jacobson:2011}, optimzied centers of rotations~\cite{Le:2016}, etc. 
Another line of works explicitly model pose specific skin deformation from sculpted or captured example poses.
For example, pose space deformation (PSD)~\cite{Lewis:2000} uses radial basis functions to interpolate between artist-sculpted surface deformations,
\cite{Kurihara:2004} extends PSD to weighted PSD, and 
\cite{Allen:2002} uses $k$-nearest neighbor interpolation. 
EigenSkin~\cite{Kry:2002} constructs compact eigenbases to capture corrections to LBS learned from examples.
The SCAPE model~\cite{scape2005} decomposes pose deformation of each mesh triangle into a rigid rotation $R$ from its body part and a non-rigid deformation $Q$ and learns $Q$ as a function of nearby joints, 
and BlendSCAPE~\cite{blendscape2012} extends this expressing each triangle's rigid rotation as a linear blend of rotations from multiple parts. 
\cite{SMPL:2015} learns a statistical body model SMPL that skins the body surface from linear pose blendshapes along with identity blendshapes.
More recently, \cite{Bailey:2018} uses neural networks to approximate the non-linear component of surface mesh deformations from complex character rigs to achieve real-time deformation evaluation for film productions.
Still, skinning remains one of the most challenging problems in the animation of virtual characters;
thus, we illustrate that our approach has the capability to overcome some errors in the skinning process.

\textbf{Cloth Skinning and Capture:}
A number of authors have made a library of cloth versus pose built primarily on simulation results and pursued ways of skinning the cloth for poses not in the library. 
\cite{Wang:2010} looks up a separate wrinkle mesh for each joint and blends them, and 
similarly \cite{Xu:2014} queries nearby examples for each body region and devises a sensitivity-optimized rigging scheme to deform each example before blending them.
\cite{Kim:2013} incrementally constructs a secondary cloth motion graph.
\cite{deAguiar:2010} learns a linear function for the principal component coefficients of the cloth shape, and 
\cite{Hahn:2014} runs subspace simulation using a set of adaptive bases learned from full space simulation data. 
Extending the SCAPE model to cloth, \cite{DRAPE2012} decomposes per-triangle cloth deformation into body shape induced deformation $D$, rigid rotation $R$, and non-rigid pose induced deformation $Q$, and applies PCA on $D$ and $Q$ to reduce dimensionality.
Whereas \cite{DRAPE2012} treats the cloth as a separate mesh, \cite{Neophytou:2014} models cloth as an additional deformation of the body mesh and learns a layered model. 
More recently \cite{Pons-Moll:2017} builds a dataset of captured 4D sequences and retargets cloth deformations to new body shapes by transfering offsets from body surfaces. 
The aforementioned approaches would all likely achieve more realistic results using real-world cloth capture as in \cite{zorah2018,Pons-Moll:2017} as opposed to physical simulations.

\textbf{Networks:}
Some of the aforementioned skinning type approaches to cloth and bodies learn from examples and therefore have procedural formulas and weights which often require optimization in order to define, but here we focus primarily on methods that use neural networks in a more data-driven as opposed to procedural fashion.
While we utilize procedural methods for skinning the body mesh and subsequently finding our cloth pixel locations, we use data-driven networks to define the cloth deformations; errors in the procedural skinning are simply incorporated into the offset function used to subsequently reach the data.
Several recent works used neural networks for learning 3D surface deformations for character rigs~\cite{Bailey:2018} and cloth shapes~\cite{Dib17a,zorah2018,Yang_2018_ECCV}.
In particular, \cite{Bailey:2018,Yang_2018_ECCV} input pose parameters and output non-linear shape deformations of the skin/cloth, both using a fully connected network with a few hidden layers to predict PCA coefficients. 
\cite{Dib17a} takes input images from single or multiple views and uses a convolutional network to predict 1000 PCA coefficients.
\cite{zorah2018} takes a hybrid approach combining a statistical model for pose-based global deformation with a conditional generative adversarial network for adding details on normal maps to produce finer wrinkles.
\begin{figure*}[b]
    \centering
    \begin{subfigure}[b]{0.28\textwidth}
        \includegraphics[width=0.98\linewidth]{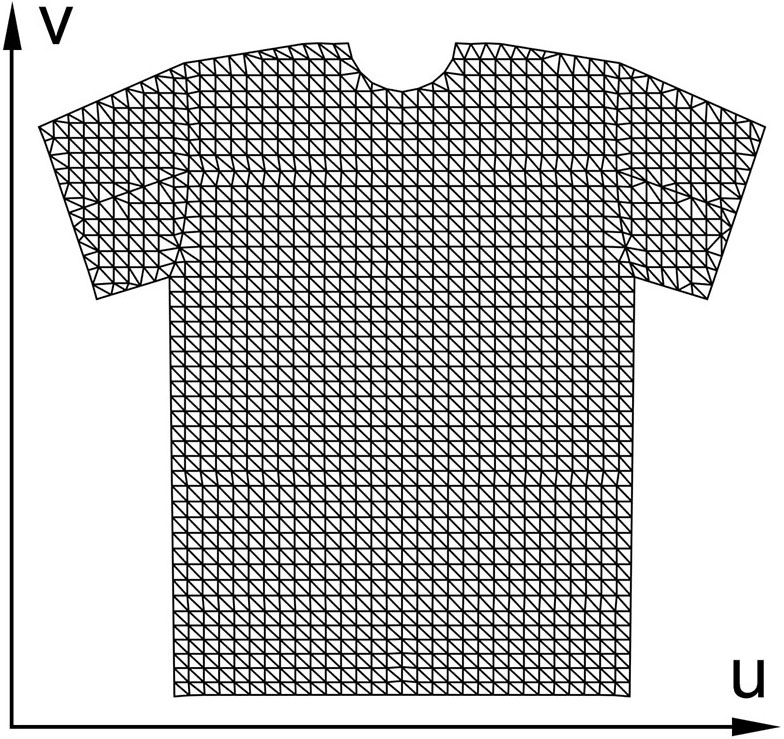}
        \caption{}
        \label{fig:front_flat_mesh}
    \end{subfigure}
    \begin{subfigure}[b]{0.34\textwidth}
        \includegraphics[width=0.98\linewidth]{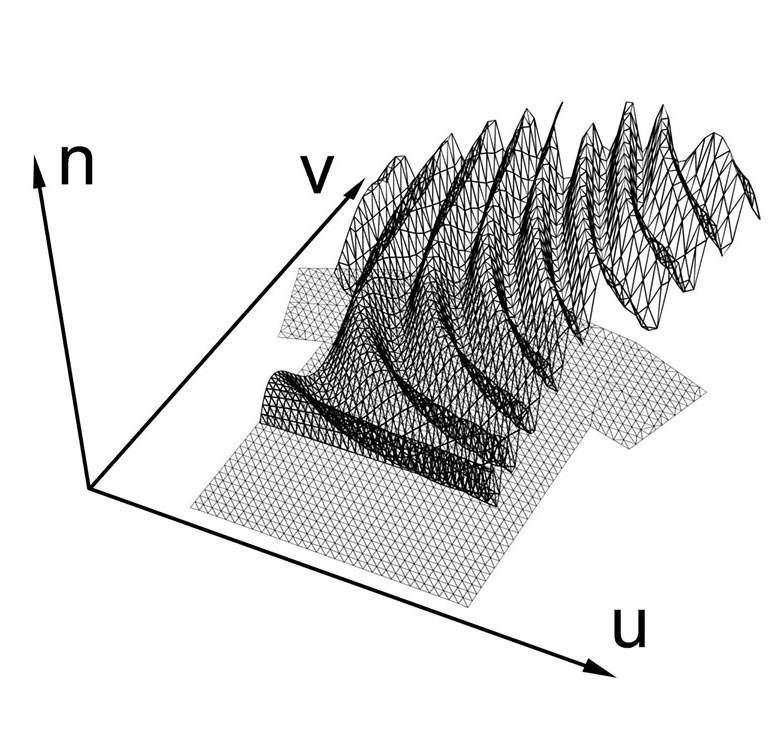}
        \caption{}
        \label{fig:front_deformed_mesh}
    \end{subfigure}
    \begin{subfigure}[b]{0.28\textwidth}
        \includegraphics[width=0.98\linewidth]{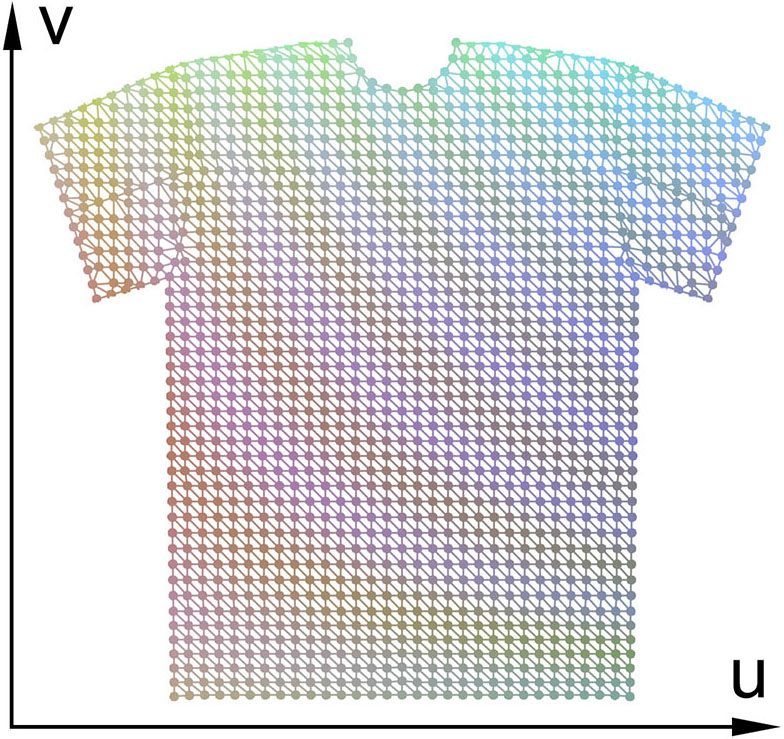}
        \caption{}
        \label{fig:front_color_deformation}
    \end{subfigure}
    \caption{Left: Triangle mesh depicted in texture space using the vertices' UV coordinates. Middle: depiction of the displacement via $(u, v, 0) + \dx$ for each vertex. Right: visualization of the displacement field $\dx$ converted into RGB values normalized to the visible $[0, 255]$ range and plotted at each vertex.}
    \label{fig:deformation_image}
\end{figure*}

\textbf{Faces:}
Face deformations bear some similarities to body skinning except there are only two bones with a single joint connecting the skull and the jaw, and most of the parameters govern shape/expression. 
We briefly mention the review paper on blendshapes~\cite{Lewis2014PracticeAT} and refer the reader to that literature for more discussions. 
However, the recently proposed \cite{FengWSWZ18} has some similarities with our approach.
They use texture coordinates similar to ours, except that they store the full 3D positions as RGB values, whereas our cloth pixels derive their 3D positions from the surface of a skinned body mesh while storing offsets from these 3D positions as RGB values. 
Extending our approach to faces, our pixels would follow the shape of the skinned face as the jaw opens and closes.
The RGB values that we would store for the face would only contain the offsets from the skinned cranium and jaw due to blendshaped expressions. 
We would not need to learn the face neutral (identity) shape or the skinning, and the offset function would simply be identically zero when no further expressions were applied, reducing the demands on the network.
Essentially, their method is what computational mechanics refers to as ``Eulerian'' where the computational domain is fixed, as opposed to a ``Lagrangian'' method with a computational data structure that follows the motion of the material (\eg, using particles).
Our approach could be considered an Arbitrary Lagrangian-Eulerian (ALE \cite{Margolin:1997}) method where the computational domain follows the material partially but not fully, \ie, our cloth pixels follow only the deformation captured by body skinning. 

\section{Pixel-Based Cloth}\label{sec:pixel_based_cloth}
We start by creating a texture map for the cloth mesh, assigning planar UV coordinates to each vertex.
For illustration, we take the front side of a T-shirt mesh as an example, see Figure~\ref{fig:front_flat_mesh}. 
Using UV space as the domain, each vertex stores a vector-valued function of displacements $\dx(u,v)= (\Delta u, \Delta v, \Delta n)$ representing perturbations in the texture coordinate and normal directions.
This can be visualized by moving each vertex by $\dx$, see Figure~\ref{fig:front_deformed_mesh}. 
These displacements can be conveniently interpreted as RGB colors stored at vertex locations in this pattern space; thus, we will refer to these vertices as \textit{cloth pixels}, see Figure~\ref{fig:front_color_deformation}.
Note that the RGB colors of the cloth pixels may contain values not in the visible range using HD image formats, floating point representations, etc. 
This framework allows us to leverage  standard texture mapping \cite{Blinn:1976,Catmull:1974,texture86} as well as other common approaches, such as using bump maps \cite{Blinn:1978} to perturb normal directions and displacement maps \cite{Cook:1984} to alter vertex positions;
these techniques have been well-established over the years and have efficient implementations on graphics hardware enabling us to hijack and take advantage of the GPU-supported pipeline for optimized performance.

\section{Cloth Images}\label{sec:cloth_images}
\begin{figure}[t]
    \centering
    \includegraphics[width=0.52\linewidth]{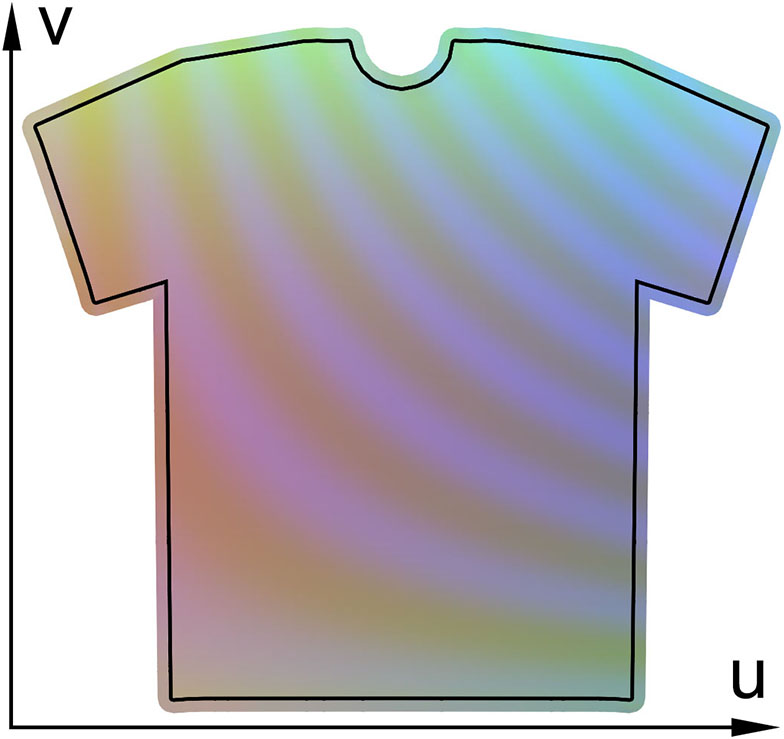}
    \caption{Standard uniform Cartesian grid of pixels for our cloth image. We add some padding to ensure smoothness on the boundaries for convolutional filters.}
    \label{fig:padded_boundary}
    \vspace{-10pt}
\end{figure}

As can be seen in Figure~\ref{fig:front_flat_mesh} and \ref{fig:front_color_deformation}, the cloth pixels are located at vertex positions and are connected via a triangle mesh topology.
CNNs exploit spatial coherency and such methods can be applied here using graph learning techniques \cite{BronsteinBLSV16,bruna2014,Defferrard:2016,HenaffBL15,Masci:2015}, see in particular \cite{Tan_2018_CVPR}.
Alternatively, since our cloth pixels have fixed UV coordinates in the two dimensional pattern space, we may readily interpolate to a uniform background Cartesian grid of pixels using standard triangle rasterization (\cite{Foley:1995}) with some added padding at the boundaries to ensure smoothness (see Figure~\ref{fig:padded_boundary}), thus facilitating more efficient application of standard CNN technologies especially via GPUs.

Note that we convert all our training data into pixel-based cloth images and train on those images directly, so that the networks learn to predict 2D images, not 3D cloth shapes. 
If one wanted to connect animation parameters to cloth vertex positions in a more fully end-to-end manner, then the interpolatory approach back and forth between the triangle mesh vertices and the pixels on the Cartesian grid would potentially require further scrutiny.
For example, the fluid dynamics community takes great care in addressing the copying back and forth of data between particle-based data structures (similar to our cloth pixels in Figure~\ref{fig:front_color_deformation}) and background grid degrees of freedom (similar to our cloth image in Figure~\ref{fig:padded_boundary}). 
Most notable are the discussions on PIC/FLIP, see \eg \cite{Jiang:2015}.

\begin{figure}[b]
    \centering
  \includegraphics[width=\linewidth]{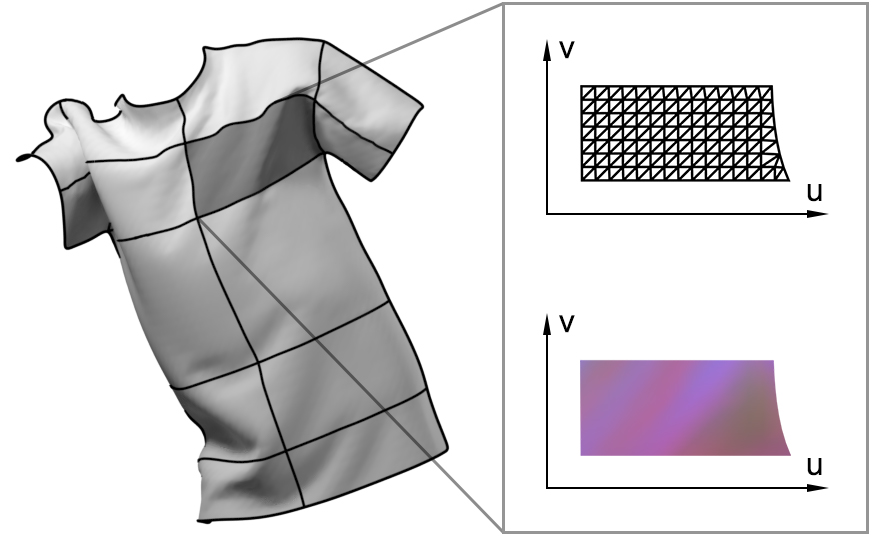}
    \caption{Left: front side of a T-shirt mesh divided into patches by a ``cage'' (depicted as black edges). Right: the triangulated cloth pixels and corresponding RGB cloth image for the highlighted patch.}
    \label{fig:cage}
\end{figure}

Quite often one needs to down-sample images, which creates problems for learning high frequency details.
Instead, we use a support ``cage'' to divide the cloth mesh into smaller patches to aid the learning process, see Figure~\ref{fig:cage}.
This notion of a cage and patch based cloth is quite powerful and is useful for capture, design, simulation, blendshape systems, etc. (see Appendix~\ref{sec:suppl_cage_patch} for more discussions).
While cloth already exhibits spatially invariant physical properties making it suitable for convolutional filters and other spatially coherent approaches, further dividing it into semantically coherent individual patches allows a network to enjoy a higher level of specialization and performance.
The only caveat is that one needs to take care to maintain smoothness and consistency across patch boundaries, but this can be achieved using a variety of techniques such as GANs~\cite{gans2014,Li_2017_CVPR}, image inpainting~\cite{Bertalmio:2000,Yu_2018_CVPR}, PCA filtering, etc.

\section{Skinning Cloth Pixels}\label{sec:skinning_cloth_pixels}
\begin{figure}[b]
    \centering
    \includegraphics[width=\linewidth]{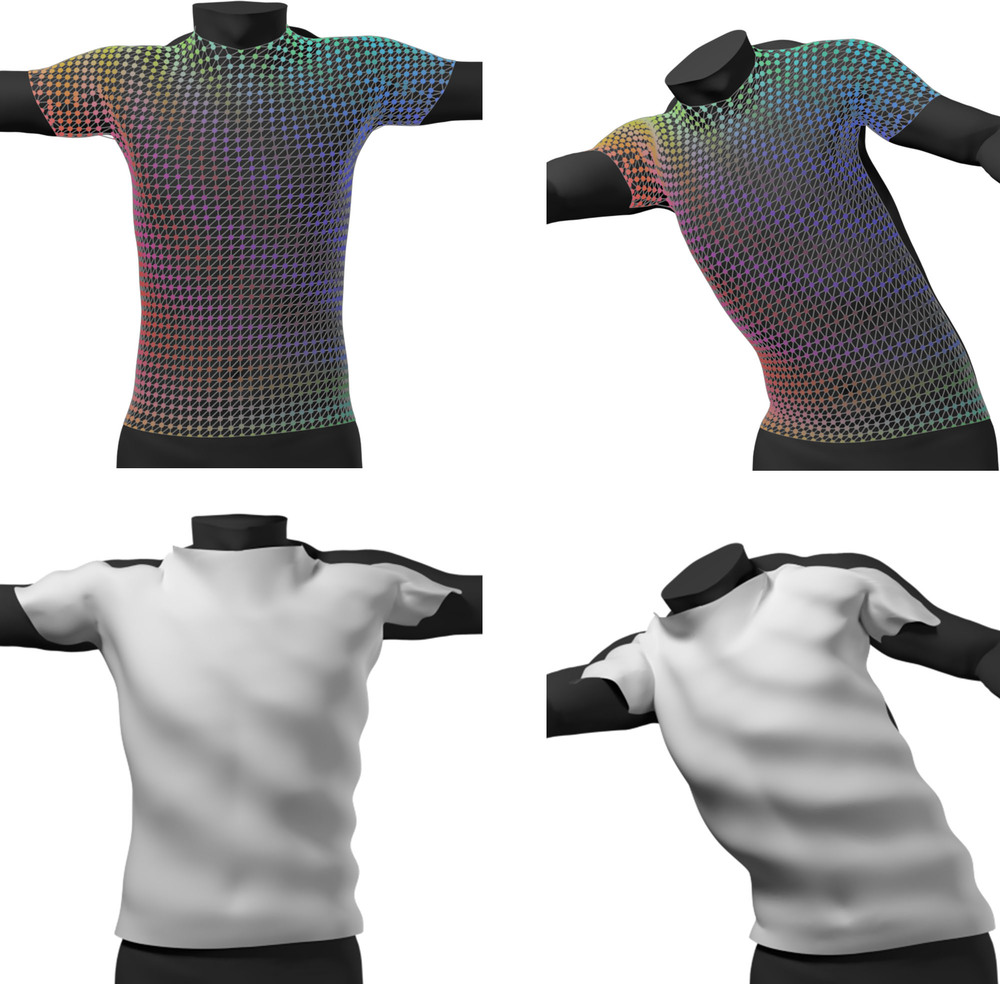}
    \caption{Top: the cloth pixels are shown embedded into body triangles with RGB values copied over from Figure~\ref{fig:front_color_deformation} in the rest pose (top left) and a different pose (top right). Bottom: The final cloth shapes obtained by adding displacements $\dx$ depicted in Figure~\ref{fig:front_deformed_mesh} to the cloth pixel locations in the top row.}
    \label{fig:tpose_skinning}
\end{figure}

\begin{figure}[t]
    \centering
    \begin{subfigure}[b]{0.3\linewidth}
        \centering
        \includegraphics[width=0.95\linewidth,trim={4cm 0.7cm 1.5cm 0.8cm},clip]{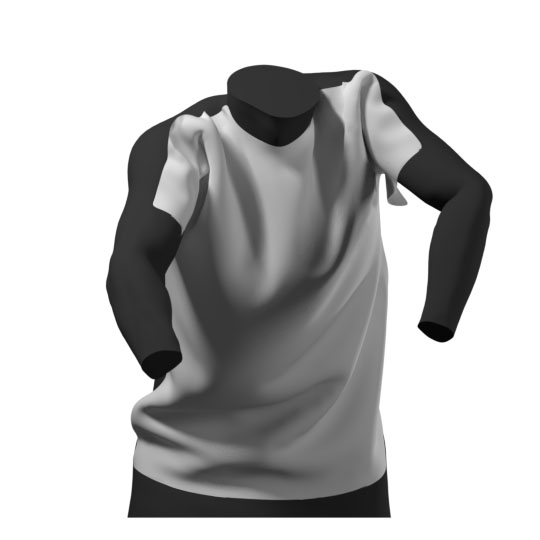}
    \end{subfigure}
    \begin{subfigure}[b]{0.33\linewidth}
        \centering
        \includegraphics[width=\linewidth,trim={0.8cm 0cm 0cm 0cm},clip]{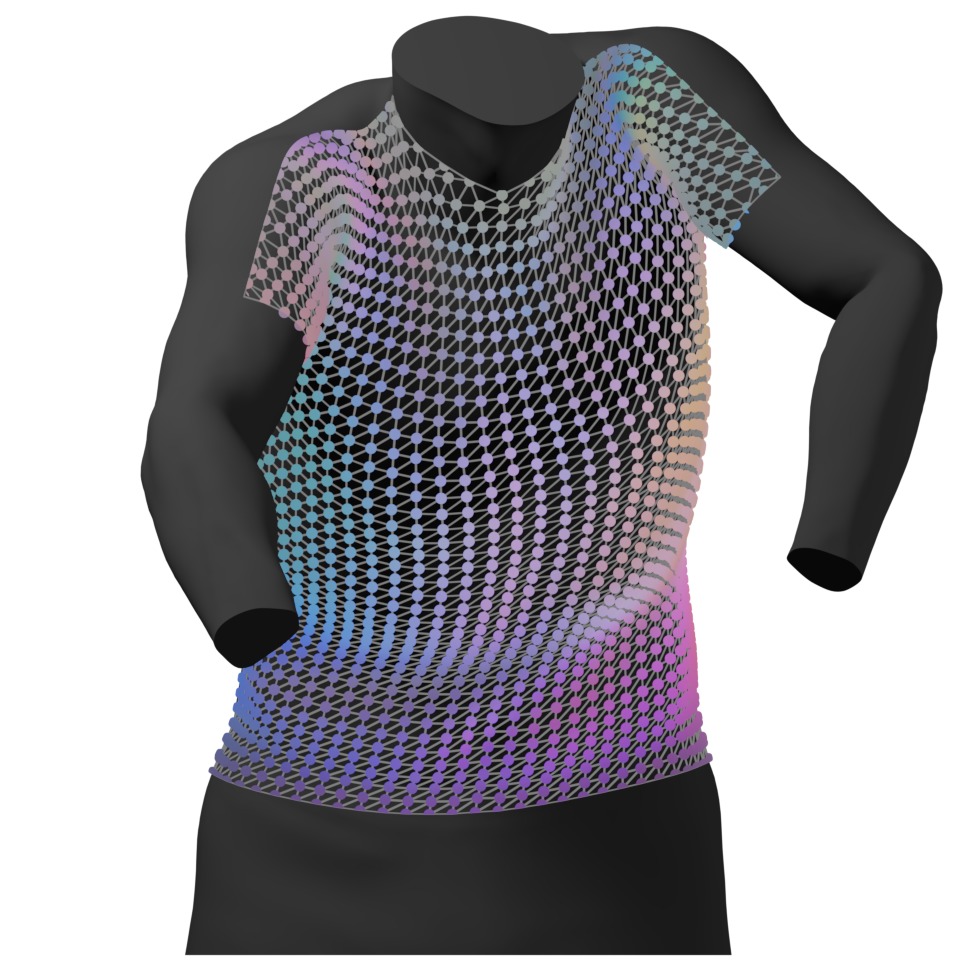}
    \end{subfigure}
    \begin{subfigure}[b]{0.35\linewidth}
        \centering
        \includegraphics[width=\linewidth]{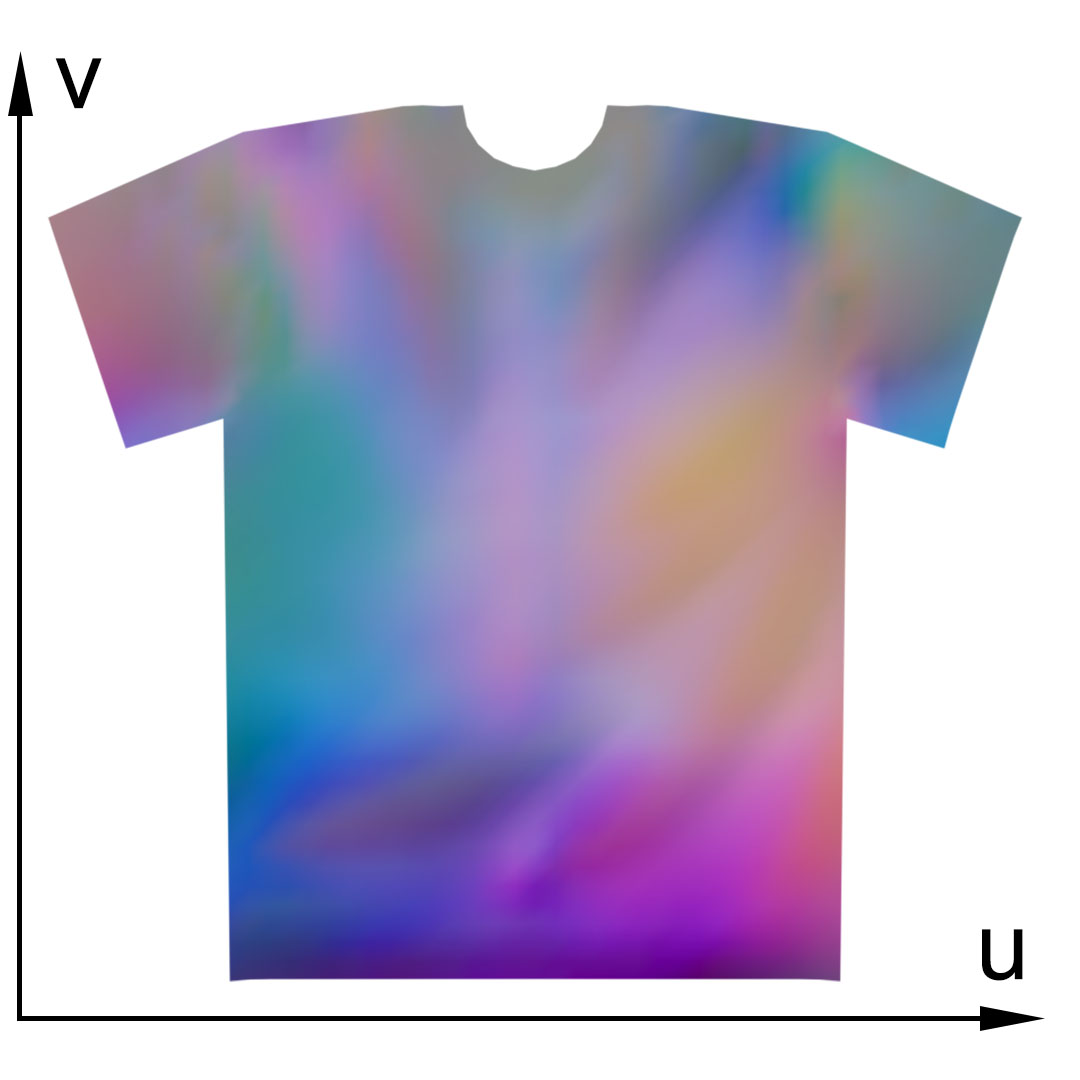}
    \end{subfigure}
    \caption{Left: part of a 3D cloth shape. Middle: cloth pixels embedded on the body mesh storing displacements $\dx$ as RGB values. Right: corresponding cloth image in the two dimensional pattern space.}
    \label{fig:create_offset_img}
\end{figure}

While the cloth pixels have fixed UV locations in their 2D pattern space, their real-world 3D positions change as the body moves.
We generate real-world positions for the cloth pixels by barycentrically embedding each of them into a triangle of the body mesh.
Then as the body mesh deforms, the real-world locations of the cloth pixels move along with the triangles they were embedded into.
Figure~\ref{fig:tpose_skinning} top row shows the pixel RGB values from Figure~\ref{fig:front_color_deformation} embedded into the rest pose and a different pose.
Applying the $\dx$ offsets depicted in Figure~\ref{fig:front_deformed_mesh} to the real-world pixel locations in Figure~\ref{fig:tpose_skinning} top row yields the cloth shapes shown in Figure~\ref{fig:tpose_skinning} bottom row.
In Figure~\ref{fig:create_offset_img}, we show the process reversed where the cloth shape shown in Figure~\ref{fig:create_offset_img} left is recorded as $\dx$ displacements and stored as RGB values on the cloth pixels embedded in the body mesh, see Figure~\ref{fig:create_offset_img} middle.
These pixel RGB values in turn correspond to a cloth image in the pattern space, see Figure~\ref{fig:create_offset_img} right.

In order to obtain barycentric embeddings of the cloth pixels to the triangles of the body mesh, we start in a rest pose and uniformly shrink the edges of the cloth mesh making it skin-tight on the body.
Since this preprocessing step is only done once, and moreover can be accomplished on a template mesh, we take some care in order to achieve a good sampling distribution of the body deformations that drive our cloth image.
Note that our formulation readily allows for more complex clothing (such as shirts/jacket collars) to be embedded on the body with overlapping folds in a non-one-to-one manner, \ie, the inverse mapping from the body texture coordinates to the cloth texture coordinates does not need to exist.
See Appendix~\ref{sec:suppl_cloth_body_texture} for more details.

\vspace*{0.5em}
One might alternatively skin the cloth as discussed in Section~\ref{sec:related_works} to obtain a candidate cloth shape, and embed our cloth pixels into the real-world skinned cloth shape, learning offsets from the skinned cloth to the simulated or captured cloth. 
The difficulty with this approach is that much of the example-based cloth can behave in quite unpredictable ways making it difficult for a network to learn the offset functions.
Thus we prefer to embed our pixels into the body geometry which deforms in a more predictable and smooth manner.
Moreover, this allows us to leverage a large body of work on body skinning as opposed to the much smaller number of works that consider cloth skinning. 

An interesting idea would be to learn the cloth shape in a hierarchical fashion, first obtaining some low resolution/frequency cloth as offsets from the human body using our image-based cloth, and then embedding pixels in that resulting cloth mesh, subsequently learning offsets from it for higher resolution.
We instead prefer analyzing the result from our image based cloth using a number of techniques including compression~\cite{Jin:2017} to see where it might require further augmentation via for example data based wrinkle maps.
That is, we do not feel that the same exact approach should be applied at each level of the hierarchy, instead preferring more specialized approaches at each level using domain knowledge of the interesting features as well as the ability to incorporate them. 
\vspace*{10pt}
\section{Network Considerations}\label{sec:networks}
\begin{figure}[h]
    \centering
    \includegraphics[width=0.6\linewidth]{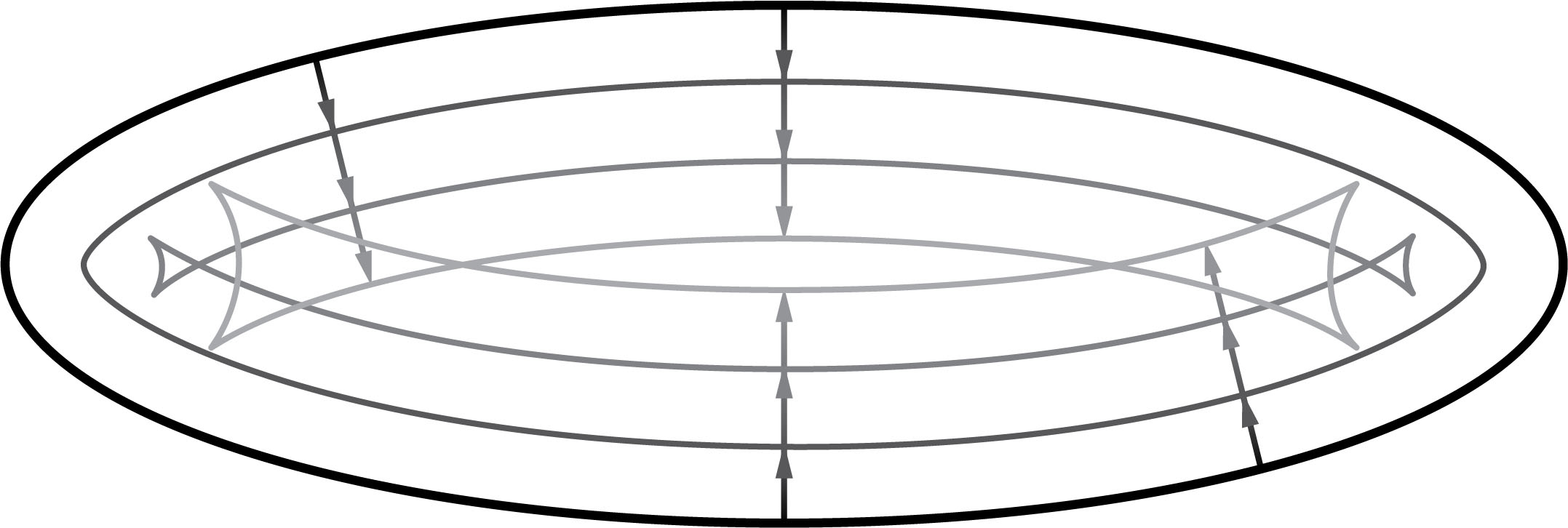}
    \caption[Caption for LOF]{An ellipse with simple constant function offsets in the normal direction, for three different constant values. (well-known swallowtail structure\protect\footnotemark{})}
    \label{fig:swallowtail}
\end{figure}

Given input pose parameters, we predict cloth images on the Cartesian grid of pixels in the pattern space. 
These images represent offsets $\dx$ from the pixels embedded to follow the body as opposed to global coordinates so that one does not need to learn what can be procedurally incorporated via body skinning (as discussed in Section~\ref{sec:related_works} in regards to faces).
Moreover, $\dx$ is parameterized in local geodesic coordinates $u$ and $v$ as well as the normal direction $n$ in order to enable the representation of complex surfaces via simpler functions, \eg, see Figure~\ref{fig:swallowtail}; even small perturbations in offset directions can lead to interesting structures.
\footnotetext{See for example page 21 of \cite{sethian1999level}.}

Although fully connected networks have been a common choice for generating dense per-vertex 3D predictions such as in \cite{Bailey:2018,Yang_2018_ECCV}, coalescing a 2D triangulated surface into a 1D vector forgos potentially important spatial adjacency information and may lead to a bigger network size as pointed out in \cite{FengWSWZ18}.
A commonly employed remedy resorts to linear dimensionality reduction methods such as PCA to recover some amount of spatial coherency and smoothness in the output, as the regularized network predicts a small number of PCA coefficients instead of the full degrees of freedom. 
Alternatively, our pixel-based cloth framework leverages convolutional networks that are particularly well-suited for and have demonstrated promising results in tasks in the image domain where the filters can share weights and exploit spatial coherency; 
our convolutional decoder takes a 1D vector of pose parameters and gradually upsamples it to the target resolution via transpose convolution operations.
As a side note, in Appendix~\ref{subsec:suppl_fc}, we illustrate that our cloth pixel framework offset functions can be approximated via a lower dimensional PCA basis, which is amenable to training and prediction via a fully connected network.

Our base loss is defined on the standard Cartesian grid image pixels weighted by a Boolean mask of the padded UV map.
One can use different norms for this loss, and empirically we find that while $L_1$ leads to slightly better quantitative metrics than $L_2$, their visual qualities are rougly similar.
Noting that normal vectors are important in capturing surface details, we experiment with incorporating an additional loss term on the per-vertex normals. 
\section{Experiments}\label{sec:experiments}
\begin{figure}[t]
    \centering
    \includegraphics[width=0.87\linewidth]{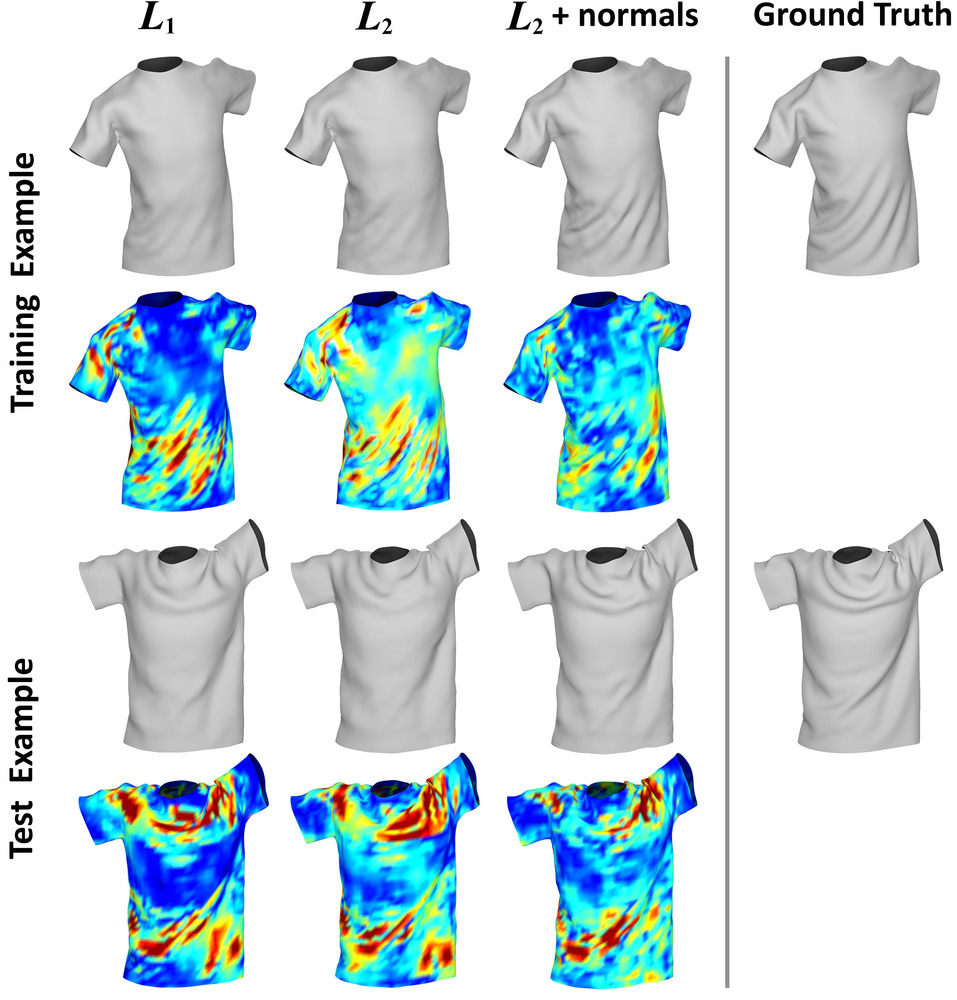}
    \caption{Network predictions/errors (blue $=$ 0, red $\geqslant$ \SI{1}{cm}) from models trained with different loss functions. While $L_1$ and $L_2$ loss on the pixels behave similarly, adding a loss term on the normals yields better visual quality.  From left to right: $L_1$ on the pixels; $L_2$ on the pixels; $L_2$ on the pixels and cosine on the normals; ground truth.}
    \label{fig:model_comp}
    \vspace{-15pt}
\end{figure} 
\textbf{Dataset Generation:} %
For the T-shirt examples, we generate 20K poses for the upper body by independently sampling rotation angles along each axis for the 10 joints from a uniformly random distribution in their natural range of motion, and then applying a simple pruning procedure to remove invalid poses, \eg, with severe nonphysical self-penetrations. 
We divided the dataset into a training set (16k samples), a regularization set (2K samples to prevent the network from overfitting), and a test set (2K samples that the optimization never sees during training). The test set is used for model comparisons in terms of loss functions and network architectures, and serves as a proxy for generalization error.
See Appendix~\ref{sec:suppl_dataset_gen} for more details.

\begin{figure}[b]
    \centering
    \includegraphics[width=0.87\linewidth]{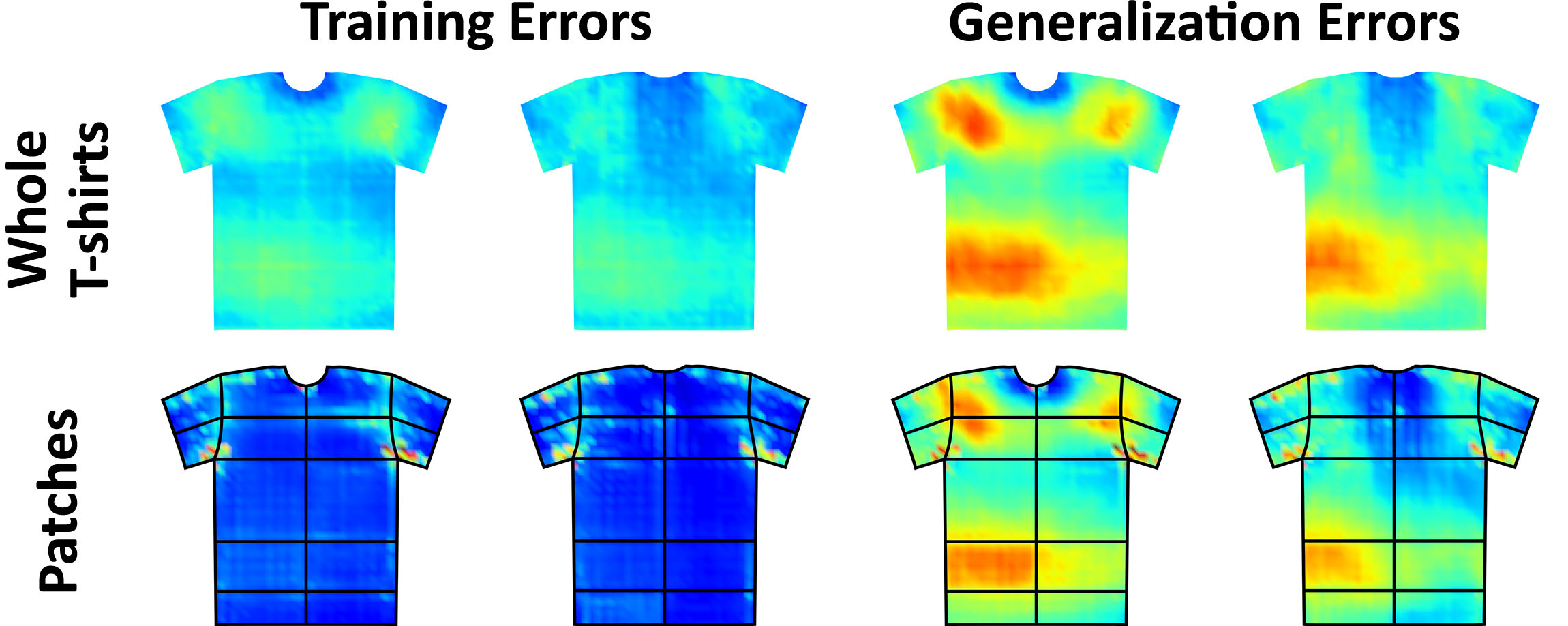}
    \caption{Dataset average per cloth pixel errors on the front/back side of the T-shirt. Top row: model trained on whole T-shirts (training/generalization error is \SI{0.37}{\cm}/\SI{0.51}{\cm}). Bottom row: models trained on patches (training/generalization error is \SI{0.20}{\cm}/\SI{0.46}{\cm}).}
    \label{fig:error_dist}
\end{figure}
\textbf{Architecture, Training, and Evaluation:} %
\begin{figure}[t]
    \centering
    \includegraphics[width=0.9\linewidth]{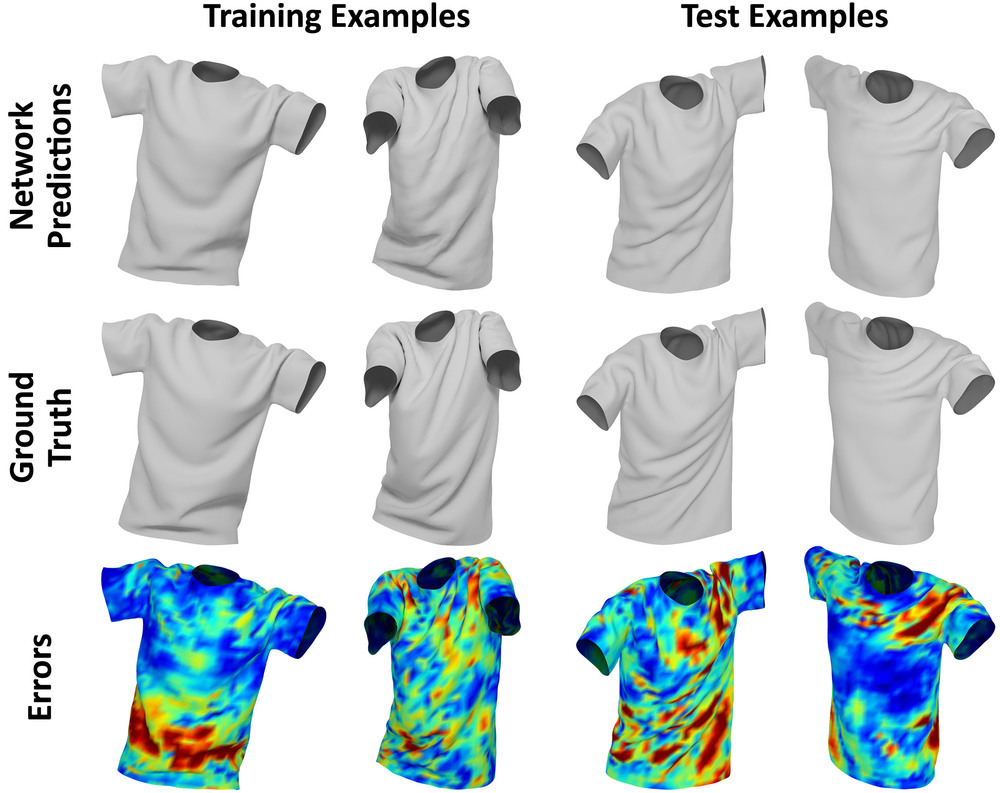}
    \caption{Network predictions and errors on training set and test set examples using our best loss model.}
    \label{fig:more_results}
    \vspace{-10pt}
\end{figure} 
\begin{figure}[b]
    \vspace{-10pt}
    \centering
    \includegraphics[width=0.8\linewidth]{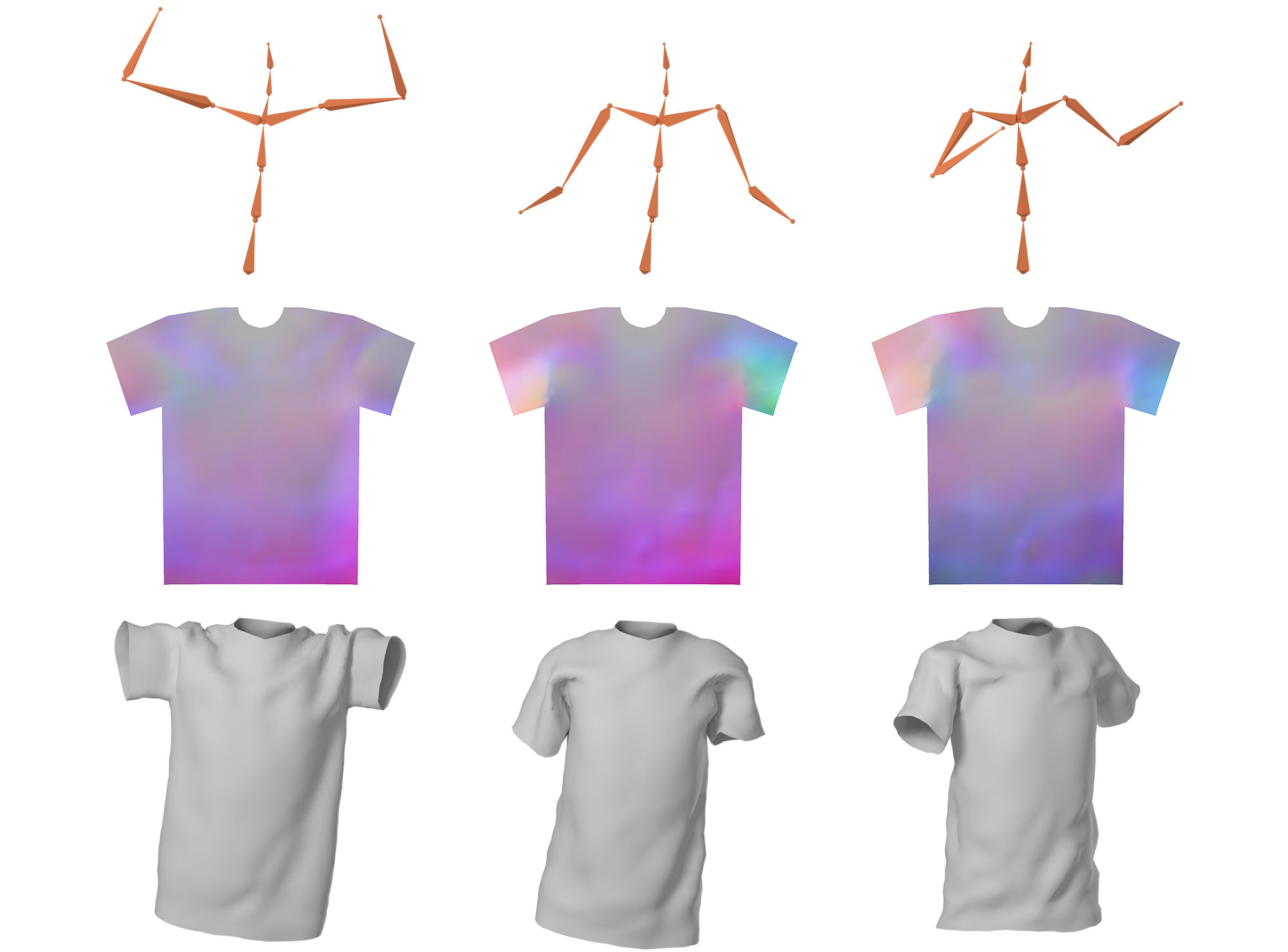}
    \caption{Evaluation on motion capture. Top: skeletal poses. Middle: predicted cloth images. Bottom: predicted cloth shapes.}
    \label{fig:mocap_pred}
\end{figure}
Our convolutional decoder network takes in $1\times1\times90$ dimensional input rotation matrices, and applies transpose convolution, batch normalization, and ReLU activation until the target output size of $256\times256\times6$ is reached, where 3 output channels represent offset values for the front side of the T-shirt and 3 channels represent those of the back.
The models are trained using the Adam optimizer~\cite{adam14} with $10^{-3}$ learning rate. 
Our implementation uses the PyTorch~\cite{pytorch} platform, and the code will be made publicly available along with the dataset.
The best visual results we obtained were from models that used additional losses on the normals, see Figure~\ref{fig:model_comp} for comparison.
Figure~\ref{fig:more_results} shows more examples in various poses from both the training and the test set and their error maps using our best loss model.
It is interesting to observe that the quantitative error metrics may not directly translate to visual quality since slight visual shift of folds or wrinkles can introduce big numerical errors.
Figure~\ref{fig:error_dist} shows the average per cloth pixel model prediction errors on the training and test set.
Unsurprisingly, the biggest errors occur near the sleeve seams and around the waist, where many wrinkles and folds form as one lifts their arms or bends.
Finally, to see how well our model generalizes to new input data, we evaluated it on a motion capture sequence from~\cite{cmumocap}, see Figure~\ref{fig:mocap_pred} and accompanying video.
\begin{figure}[t]
    \centering
    \includegraphics[width=0.86\linewidth]{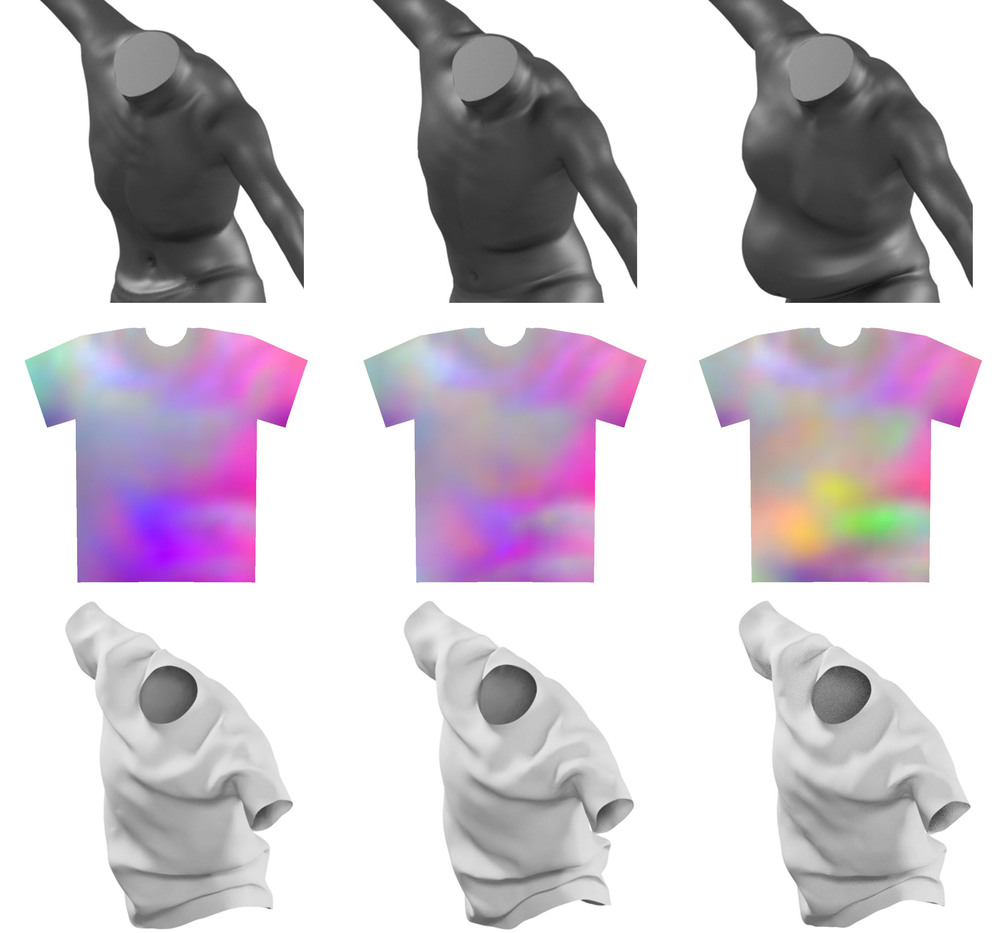}
    \caption{Training the network on unclothed body shapes that are too thin (left column) or too thick (right column) does not hinder its ability to predict cloth shapes, as compared to the ground truth (middle column). The cloth images (middle row) readily compensate for the incorrect unclothed body shape assumptions leading to similar cloth shapes (bottom row) in all three cases.}
    \vspace{-10pt}
    \label{fig:pred_diff_body_shape}
\end{figure}

\vspace{5pt}
\textbf{Modified Body Shapes:} 
The inability to obtain accurate unclothed body shapes is often seen as a real-world impediment to e-commerce clothing applications. 
Our approach embeds cloth pixels in the unclothed form and leverages procedural body skinning techniques to move those cloth pixels throughout space. 
This embedding of cloth pixels provides spatial coherency for the CNN and alleviates the need for the network to learn body shape (identity) and deformation. 
However, similar body shapes would tend to deform similarly, especially if the dimensions aren't too different. 
Thus, we intentionally modified our unclothed body shape making it too thick/thin in order to represent inaccuracies in the assumed body shape of the user. 
For each modified body shape, we use the same training data for cloth shapes noting that this merely changes the values of $\dx$ and thus the cloth image stored for each pose.  
As compared to the high variance in $\dx$ caused by folds and wrinkles, changing the body shape makes lower frequency modifications that are not too difficult for the network to learn. 
Surprisingly, the erroneously modified too thick/thin body shapes had almost no effect on the network's prediction ability indicating that our approach is robust to inaccuracies in the unclothed body shape. 
See Figure~\ref{fig:pred_diff_body_shape} and Figure~\ref{fig:pred_w_shirt} left.

\begin{figure}[t]
    \centering
    \includegraphics[width=0.86\linewidth]{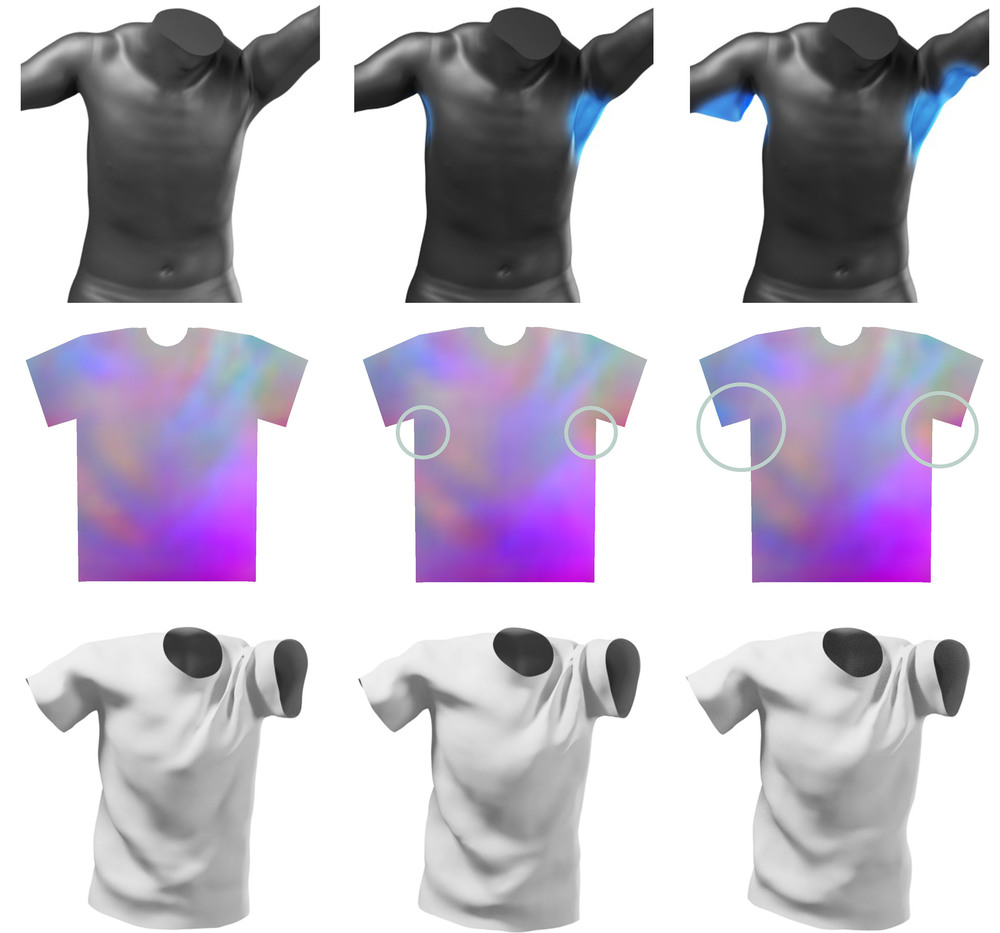}
    \caption{Training the network using a body skinning method that contains artifacts (shown in blue) does not hinder its ability to predict cloth shapes as compared to the ground truth (left column).  The cloth images (middle row) readily compensate (see regions annotated by circles) for the skinning artifacts leading to similar cloth shapes (bottom row) in all three cases.}
    \label{fig:pred_skinning_artifact}
    \vspace{-10pt}
\end{figure}
\begin{figure}[b]
    \centering
    \vspace{0pt}
    \includegraphics[width=0.8\linewidth]{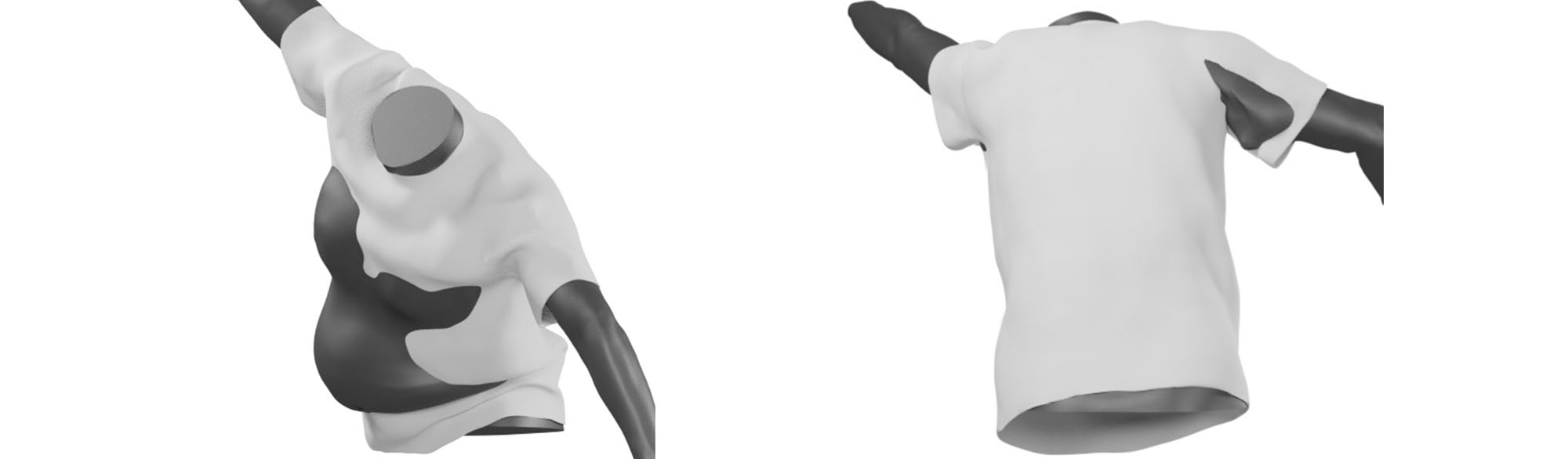}
    \caption{The CNN predicts the correct cloth shape even when the unclothed shapes are so erroneous that they penetrate the clothing. In these cases, the network merely predicts offsets in the negative normal direction. Left: dressed version of Figure~\ref{fig:pred_diff_body_shape} right. Right: dressed version of Figure~\ref{fig:pred_skinning_artifact} right.}
    \label{fig:pred_w_shirt}
\end{figure}

\textbf{Skinning Artifacts:} 
Whether using an accurate unclothed body shape or not, clearly body skinning is not a solved problem; thus, we modified our skinning scheme to intentionally create significant artifacts using erroneous bone weights.
Then, we trained the CNN as before noting that the cloth training images will be automatically modified whenever skinning artifacts appear.
The erroneous skinning artifacts had almost no effect on the network's prediction ability indicating that our approach is robust to inaccuracies in the body skinning. 
See Figure~\ref{fig:pred_skinning_artifact} and Figure~\ref{fig:pred_w_shirt} right.

\textbf{Cloth Patches:} %
\begin{figure}
    \centering
        \includegraphics[width=0.9\linewidth]{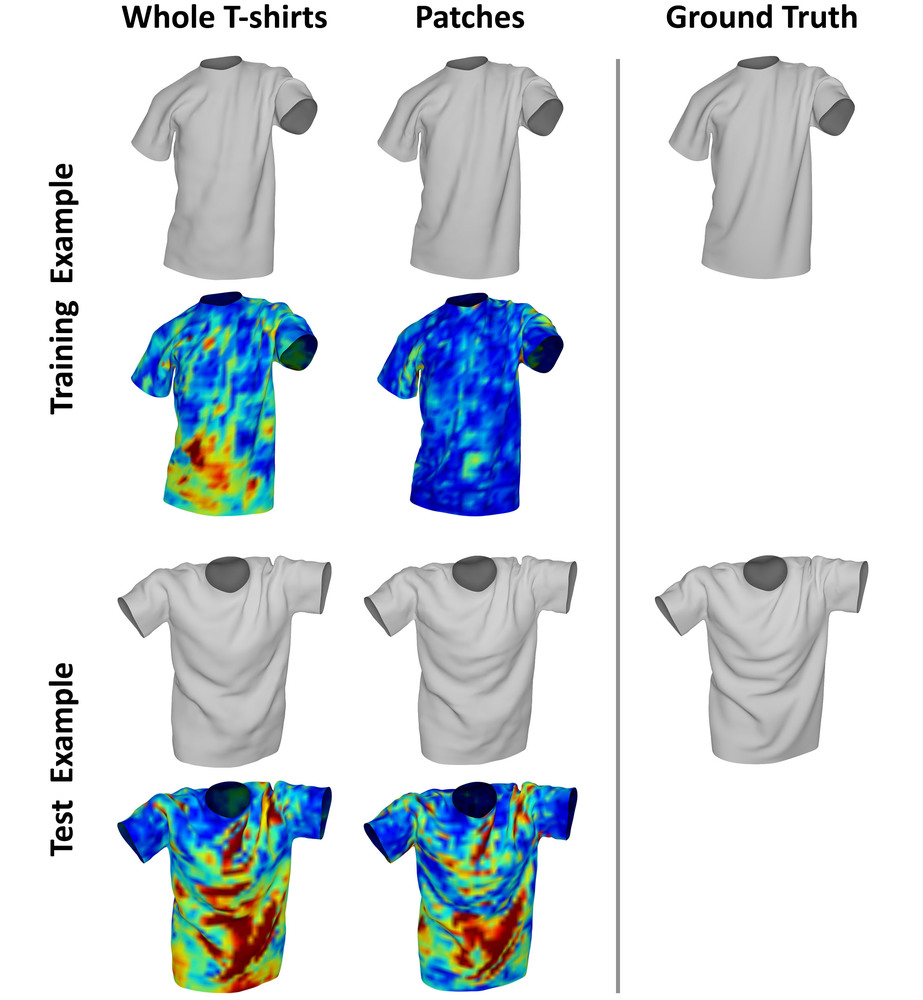}
    \caption{Comparison of network predictions/errors from model trained on whole T-shirts versus models trained on patches. The latter can better capture folds and wrinkles.}
    \label{fig:patch_comp}
    \vspace{-15pt}
\end{figure}
As mentioned in Section~\ref{sec:cloth_images}, we can segment the cloth mesh into smaller semantically coherent pieces, and then train separate networks on these individual patches to achieve better results.
Figure~\ref{fig:error_dist} shows that the models trained on the patches yield lower errors.
See Figure~\ref{fig:patch_comp} for visual comparison.
One can use a variety of methods to achieve visually continuous and smooth results across the patch boundaries.
\begin{figure}[b]
    \centering
     \vspace{-10pt}
    \includegraphics[width=0.9\linewidth]{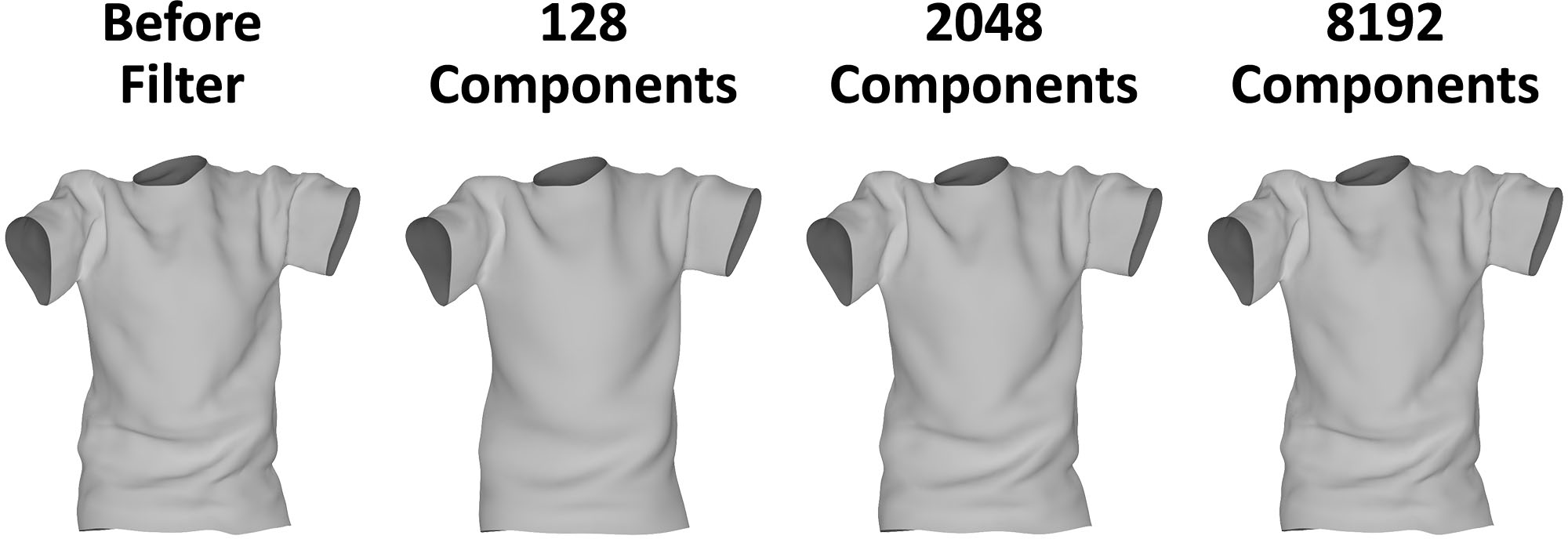}
    \caption{PCA filtering on a stitched mesh from predicted patches (an example from the test set).}
    \label{fig:pca_filtering}
\end{figure}
For example, one can precompute the PCA bases of the whole mesh on the training samples, and then project the stitched mesh onto a subset of those bases.
Since the simulation/captured data do not have kinks at patch boundaries, the PCA bases also will not have kinks at boundaries unless one gets into ultra-high frequency modes that represent noise; thus, reconstructing the network predicted results using a not too high number of PCA bases acts as a filter to remove discontinuities at patch boundaries.
In our experiments, using 2048 components leads to the best filtering results, see Figure~\ref{fig:pca_filtering}.

\begin{figure}[t]
    \centering
    \includegraphics[width=0.9\linewidth]{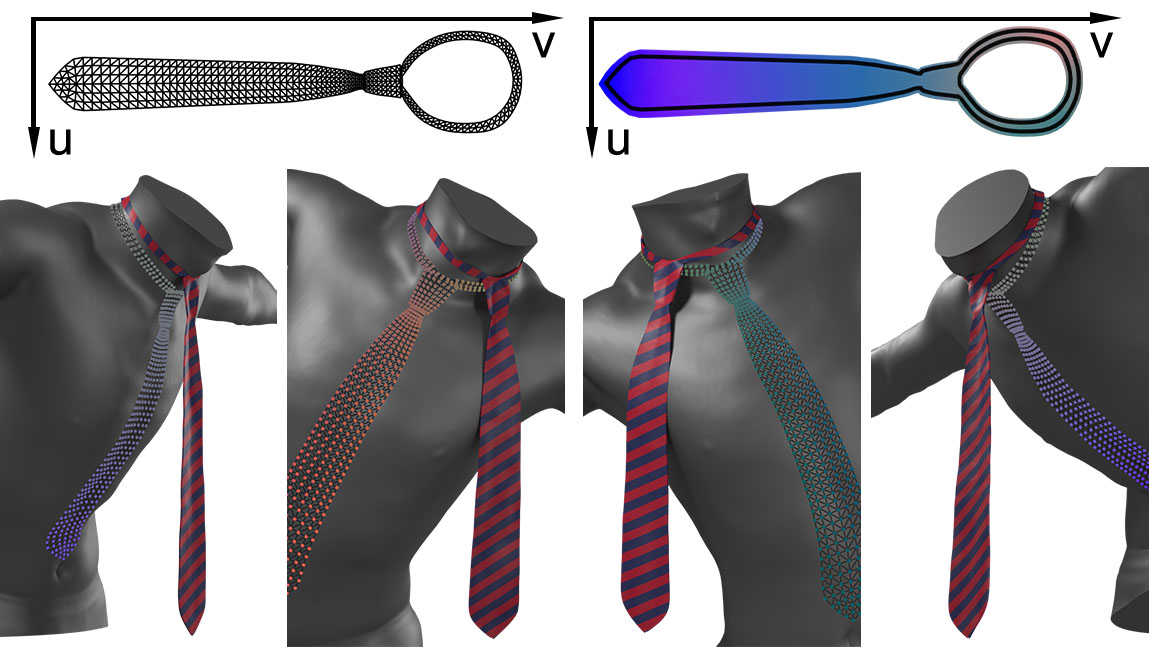}
    \caption{Top left: triangle mesh of necktie in pattern space. Top right: a necktie image. Bottom: network predictions of neckties in different poses (also, necktie pixels are shown embedded on the skinned body mesh).}
    \label{fig:necktie_result}
\end{figure}

\textbf{Necktie:} %
For generality we also show a necktie example, which unlike the T-shirt, exhibits much larger deformation as the body moves; the maximum per-vertex offset value can be over 50 centimeters.
See Figure~\ref{fig:necktie_result}, and Appendix~\ref{sec:suppl_neckties} for more details.

\section{Conclusion and Future Work}
In conclusion, we have introduced a new flexible pixel-based framework for representing virtual clothing shapes as offsets from the underlying body surface, and further illustrated that this data structure is especially amenable to learning by convolutional neural networks in the image space. 
Our preliminary experiments show promising results with CNNs successfully predicting garment shapes from input pose parameters, and we are optimistic that the results could be further improved with better and more advanced network technologies.

For future work, we would like to leverage real-world captured cloth data and generalize our approach to a larger variety of garment types and materials as well as body types.
We would also like to explore alternative network architectures, loss functions, and training schemes to enhance the visual quality of the predictions.
In addition, while our evaluation on the motion capture sequence already appears quite smooth in time, we would like to experiment with techniques such as 3D CNNs and recurrent neural networks to achieve better temporal coherency. 

\ifcvprfinal
\section*{Acknowledgements}
Research supported in part by ONR N000014-13-1-0346, ONR N00014-17-1-2174, ARL AHPCRC W911NF-07-0027, and generous gifts from Amazon and Toyota.
In addition, we would like to thank Radek Grzeszczuk for initiating conversations with Amazon and those interested in cloth there, Andrew Ng for many fruitful discussions on cloth for e-commerce, and both Reza and Behzad at ONR for supporting our efforts into machine learning.
Also, we greatly appreciate the remarkable things that Jen-Hsun Huang (Nvidia) has done for both computer graphics and machine learning; this paper in particular was motivated by and enabled by a combination of the two (and inspirations from chatting with him personally).
NJ is supported by a Stanford Graduate Fellowship, 
YZ is supported by a Stanford School of Engineering Fellowship, and 
ZG is supported by a VMWare Fellowship.
NJ would also like to personally thank a number of people who helped contribute to our broader efforts on data-driven cloth, including Davis Rempe, Haotian Zhang, Lucy Hua, Zhengping Zhou, Daniel Do, and Alice Zhao. 
\fi
\clearpage

\appendix
\renewcommand{\thesection}{\Alph{section}}
\section*{Appendices}

\section{Cloth/Body Texture Space}\label{sec:suppl_cloth_body_texture}
\begin{figure}[h]
    \centering
    \includegraphics[width=0.5\linewidth]{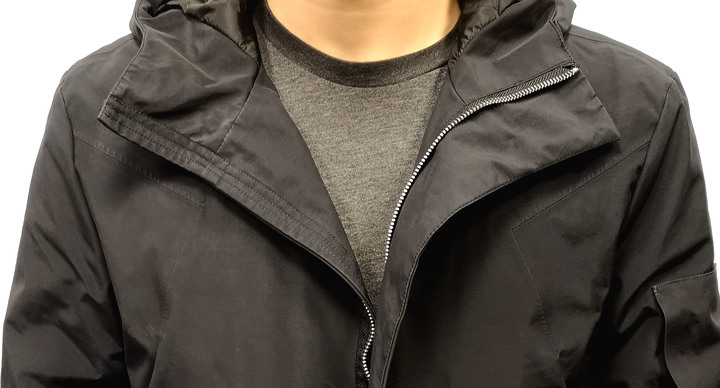}
    \caption[Caption for LOF]{Collars such as this one are more naturally associated with the chest than the neck. Our approach can handle such a non-invertible many-to-one mapping from the cloth texture space to the body texture space.}
    \label{fig:jacket_collar}
\end{figure}

It is important to note that that we do not assume a one-to-one mapping between the cloth texture coordinates and the body texture coordinates; rather, we need only a mapping from the cloth texture space to the body texture space (invertibility is not required). 
This allows for the ability to handle more complex real-life clothing such as the collars of shirts and jackets, which would naturally be embedded to the shoulder/chest areas on the body causing them to overlap with other parts of the same garment (and/or other garments). 
See for example Figure~\ref{fig:jacket_collar}.

\section{Image Editing}\label{sec:suppl_image_edit}
\begin{figure*}[b]
    \centering
    \includegraphics[width=\textwidth]{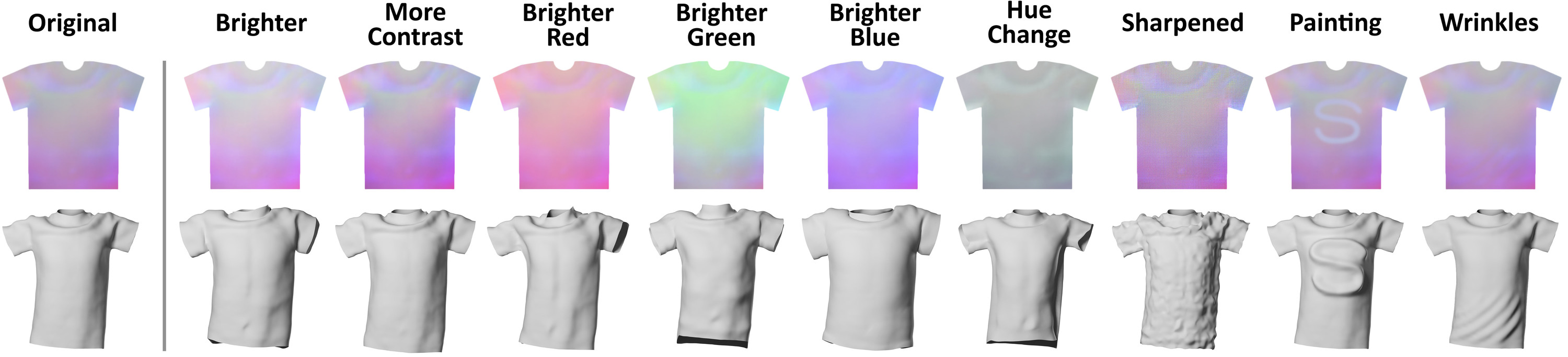}
    \caption{Various image editing operations applied to a given cloth image (top row) and their corresponding modified cloth shapes (bottom row). Note that although the wrinkle lines blended into the image in the last column are hard to see, the resulting wrinkles are clearly visible.}
    \label{fig:image_edit}
\end{figure*}
Our pixel-based cloth framework enables convenient shape modification via image editing.
Since the displacement maps represent offsets from the locations of the embedded cloth pixels on the skinned body surface, we can achieve easy and rather intuitive control by manipulating their RGB values in the image space. 
For example, adjusting the brightness of the texture coordinates channels (red and green) induces shifting of the cloth shape, whereas adjusting the normal directions channel (blue) leads to shrinking or inflation.
Moreover, one can add features to the cloth shape by painting in image space, especially using a blue brush that changes the offset values in the normal directions. 
Furthermore, one can transfer features from another cloth image by selective image blending, \eg, adding wrinkle lines. 
See Figure~\ref{fig:image_edit} for a set of modified cloth shapes resulting from image editing operations.

\section{Cage and Patch Based Cloth}\label{sec:suppl_cage_patch}
\begin{figure}[h]
    \centering
    \begin{subfigure}[b]{0.45\linewidth}
        \centering
        \includegraphics[width=0.9\linewidth]{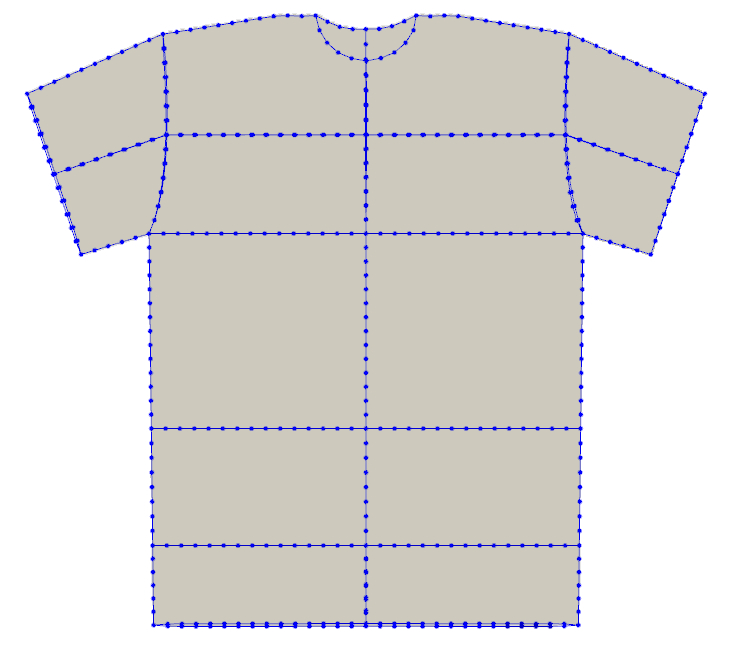}
        \caption{Cage structure defined on a T-shirt mesh.}
        \label{fig:cloth_curves}
    \end{subfigure}
    \hspace{10pt}
    \begin{subfigure}[b]{0.45\linewidth}
        \centering
        \includegraphics[width=0.9\linewidth]{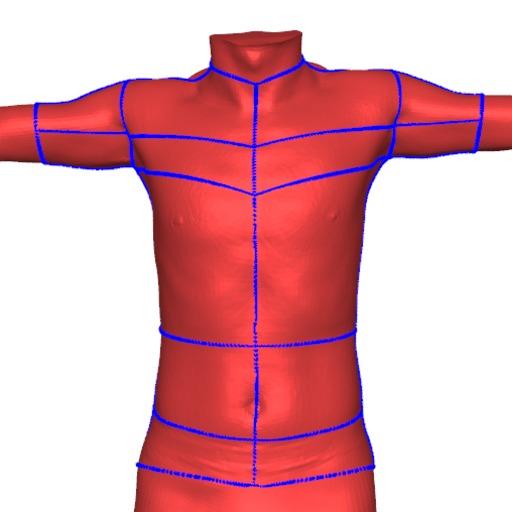}
        \caption{Corresponding cage defined on the body.}
        \label{fig:body_curves}
    \end{subfigure}
    \caption{The cage is defined on the cloth mesh and the body surface as a lower-dimensional support structure.}
    \label{fig:cloth_body_curves}
\end{figure}

Given a cloth mesh, we can create a wire ``cage'' that defines a support structure for its overall shape, \eg, by tracing its intrinsic seams, characteristic body loops (\eg, chest, waist, hip, arms), etc. 
See Figure~\ref{fig:cloth_curves}.
The cage structure conveniently divides the cloth surface into patches bound by boundary curves, and this cage and patch based computational structure affords a hierarchical data-driven framework where different specialized methods can be applied at each level.
Note that the same cage structure is also defined on the body surface to facilitate correspondences, see Figure~\ref{fig:body_curves}.

To obtain the shape of the cage when the clothing is dressed on a person, one can interpolate from a set of sparse marker/characteristic key points. 
That is, given the locations of the key points, one can reconstruct the cage.
This can be represented as a constrained optimization problem to find a smooth curve that passes through the constraint points. 
Specifically, one can interpolate the points with a piecewise cubic spline curve while attempting to preserve the geodesic lengths between each pair of neighboring points.
Alternatively, one could train a neural network to learn to recover the cage from the sparse points.

One can use the reconstructed cage as a boundary condition to fill in the surface patches using a variety of methods. 
In particular, one can build a blendshapes basis for each patch and select blendshape weights based on the shape of the boundary cage. 
A cage vertex's basis function can be computed, for example, by solving a Poisson equation on the patch interior with boundary conditions identically zero except at that vertex where the boundary condition is set to unity. 
Then, any perturbation of the cage can be carried to its interior. 
For example, given the offsets of the cage from its position on the skinned body in texture and normal coordinates, one can evaluate the basis functions to quickly compute offsets for the interior of the patch.

\begin{figure}[h]
    \centering
    \begin{subfigure}[b]{0.32\linewidth}
    \includegraphics[width=\linewidth]{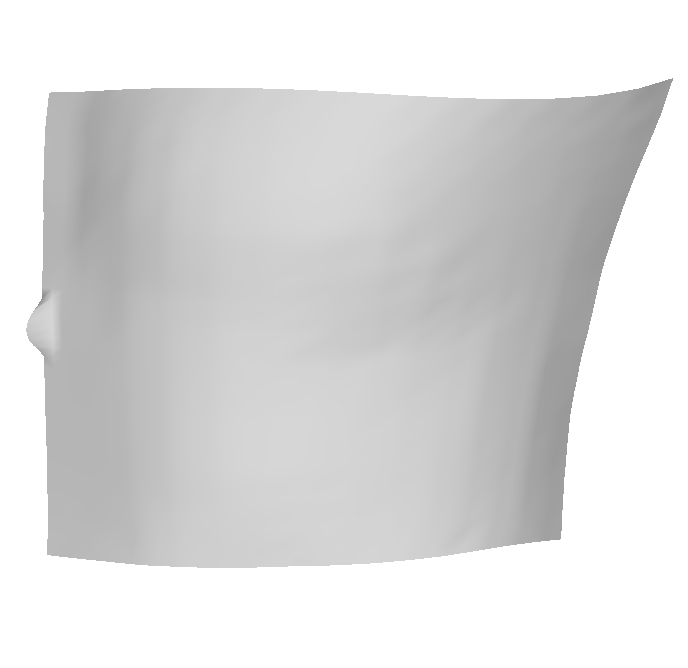}
    \caption{}
    \label{fig:one_bump_boundary}
    \end{subfigure}
    \begin{subfigure}[b]{0.32\linewidth}
    \includegraphics[width=\linewidth]{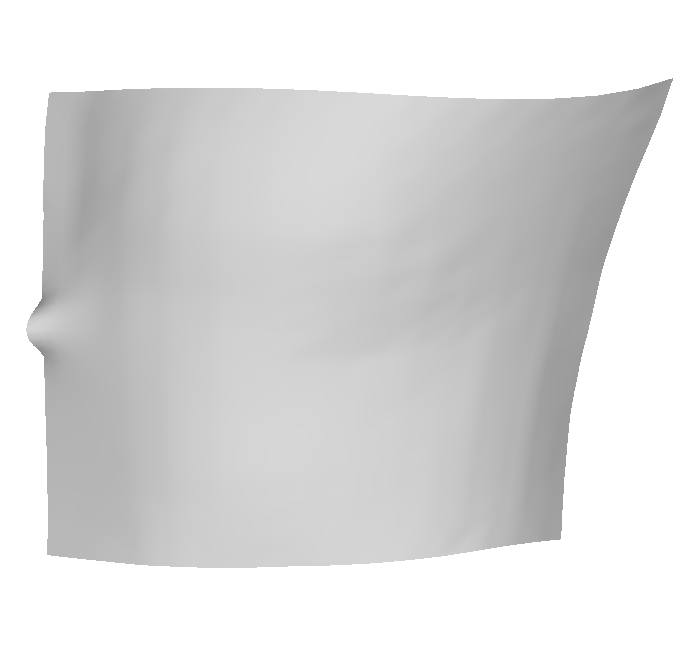}
    \caption{}
    \label{fig:one_bump_vanilla}
    \end{subfigure}
    \begin{subfigure}[b]{0.32\linewidth}
    \includegraphics[width=\linewidth]{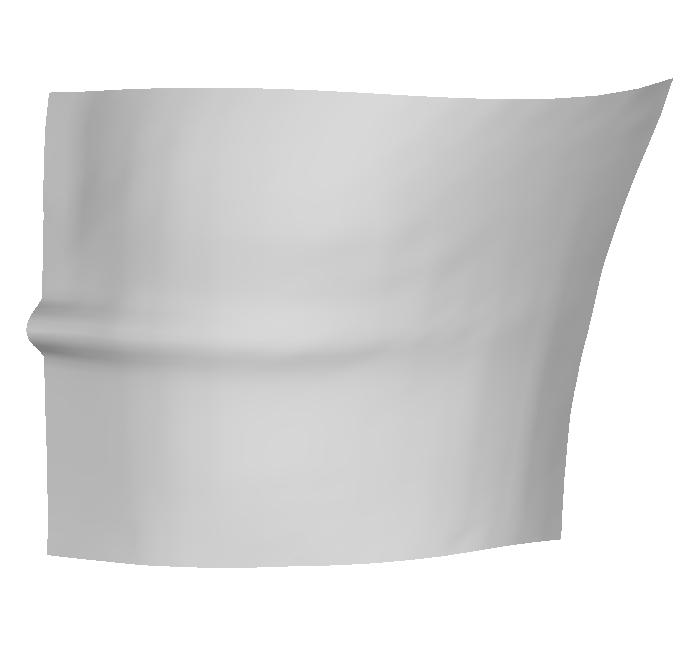}
    \caption{}
    \label{fig:one_bump_anis}
    \end{subfigure}
    \caption{An input boundary perturbation (a) can be used in a blendshape basis to obtain interior patch deformations: isotropic (b), anisotropic (c).}
    \vspace{-10pt}
    \label{fig:one_bump}
\end{figure}

\begin{figure}[h]
    \captionsetup[subfigure]{justification=centering}
    \centering
    \begin{subfigure}{0.4\linewidth}
    \centering
    \includegraphics[width=.8\linewidth]{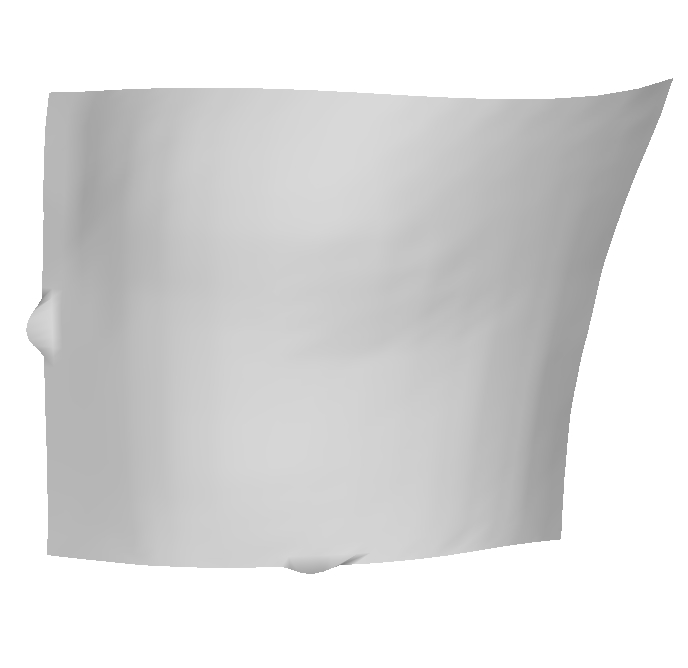}
    \caption{boundary condition}
    \label{fig:two_bump_bdry}
    \end{subfigure}
    \begin{subfigure}{0.4\linewidth}
    \centering
    \includegraphics[width=.8\linewidth]{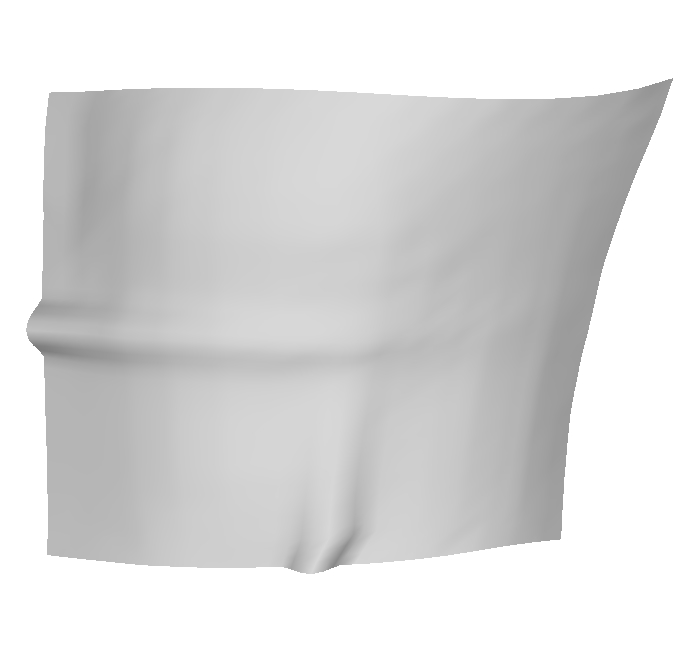}
    \caption{patch shape}
    \label{fig:two_bump_anis}
    \end{subfigure}
    \caption{Two small perturbations on the boundary yields two folds coming together.}
    \label{fig:two_bump}
\end{figure}

\begin{figure}
    \captionsetup[subfigure]{justification=centering}
    \centering
    \begin{subfigure}{0.4\linewidth}
    \centering
    \includegraphics[width=.8\linewidth]{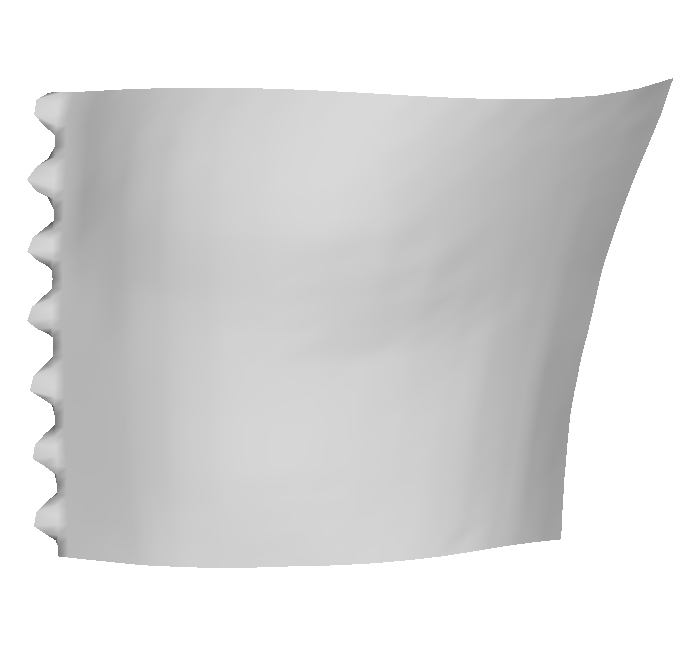}
    \caption{boundary condition}
    \label{fig:sine_wave_bdry}
    \end{subfigure}
    \begin{subfigure}{0.4\linewidth}
    \centering
    \includegraphics[width=.8\linewidth]{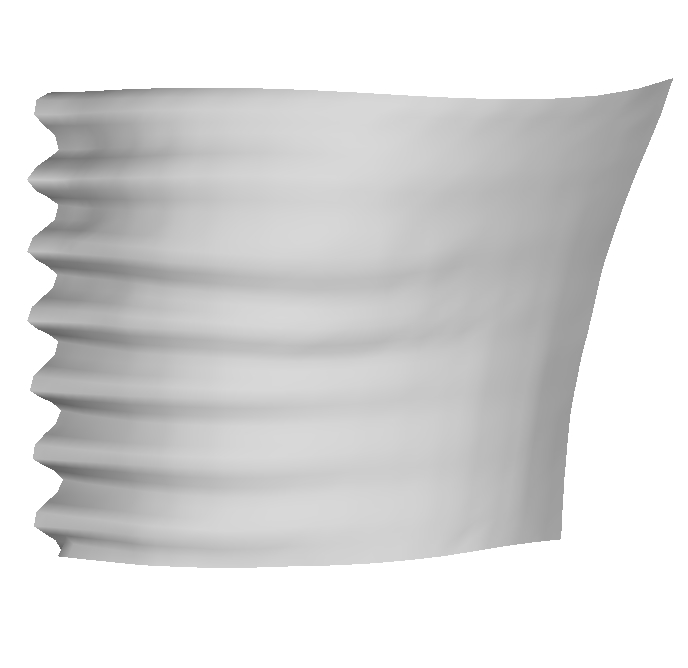}
    \caption{patch shape}
    \label{fig:sine_wave_anis}
    \end{subfigure}
    \caption{A sine wave perturbation on the boundary yields smooth wrinkles.}
    \label{fig:sine_wave}
\end{figure}

\begin{figure}
    \captionsetup[subfigure]{justification=centering}
    \centering
    \begin{subfigure}{0.4\linewidth}
    \centering
    \includegraphics[width=.8\linewidth]{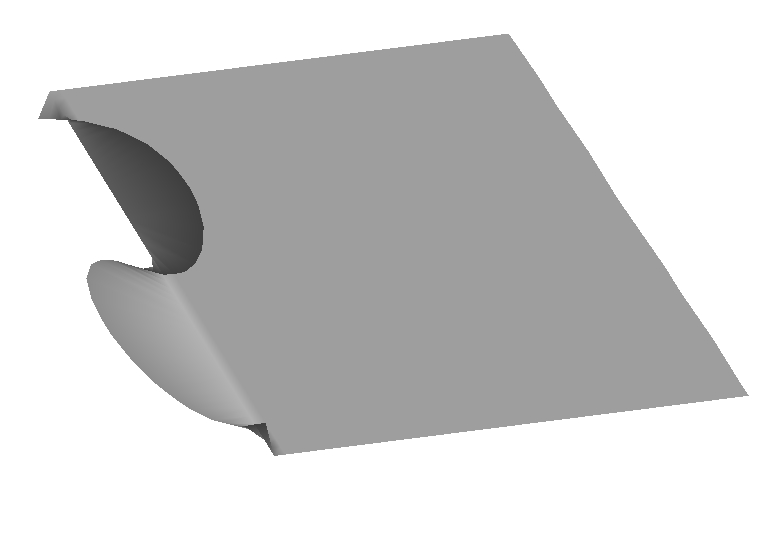}
    \caption{boundary condition}
    \label{fig:overturning_bdry}
    \end{subfigure}
    \begin{subfigure}{0.4\linewidth}
    \centering
    \includegraphics[width=.8\linewidth]{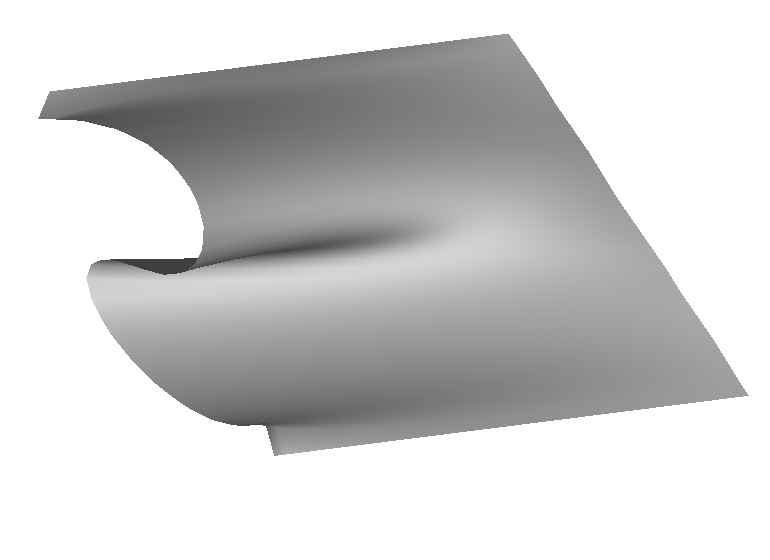}
    \caption{patch shape}
    \label{fig:overturning_anis}
    \end{subfigure}
    \caption{An S-shaped boundary yields an overturning wave shape.}
    \label{fig:overturning}
\end{figure}

For a simple illustration, the boundary perturbation in Figure~\ref{fig:one_bump_boundary} is extended to the patch interior using the Poisson equation basis functions to obtain the result shown in Figure~\ref{fig:one_bump_vanilla}. 
To achieve more interesting deformations, one could use anisotropic Poisson equations to construct the basis functions. 
Figure~\ref{fig:one_bump_anis} shows the boundary perturbation in Figure~\ref{fig:one_bump_boundary} evaluated using anisotropic basis functions. 
Also, see Figures \ref{fig:two_bump}, \ref{fig:sine_wave}, and \ref{fig:overturning}.  
One could also create basis functions via quasistatic simulations.


Moreover, one can use this cage structure as an intermediary for designing and dressing garments onto the body leveraging the correspondence to body curves shown in Figure~\ref{fig:cloth_body_curves}.

\section{Dataset Generation}\label{sec:suppl_dataset_gen}
\subsection{Body Modeling}
\begin{figure}[h]
    \centering
    \includegraphics[width=\linewidth]{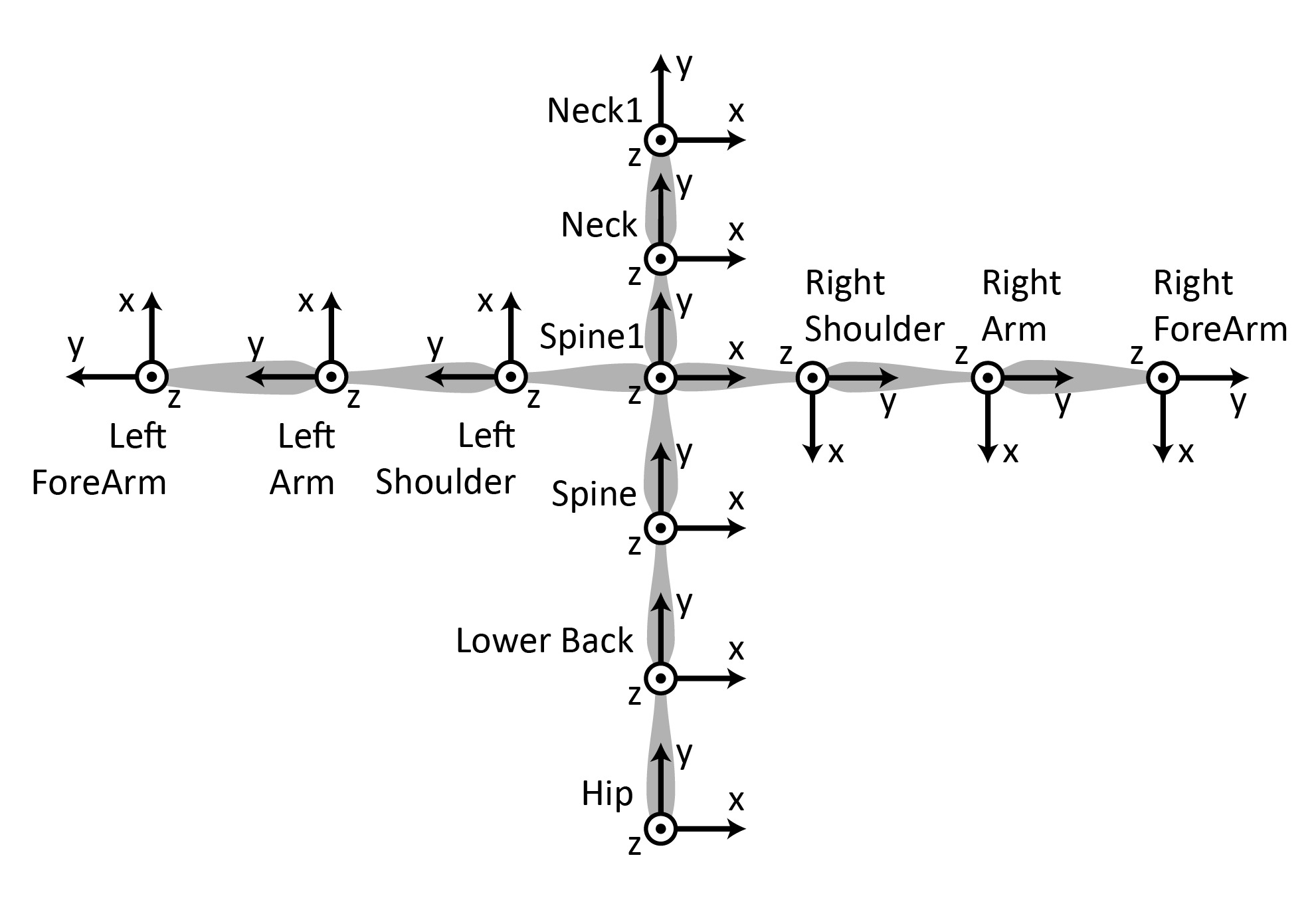}
    \caption{Skeleton structure, bone name, and axis orientation definition.}
    \label{fig:skeleton_def}
\end{figure}

We use a commercial solution\footnote{\url{https://www.artec3d.com/}} to scan a person in the T-pose. 
The initially acquired mesh is manually remeshed to a quad mesh, and then rigged to the skeleton shown in Figure~\ref{fig:skeleton_def} using Blender~\cite{blender}.

\subsection{Intentionally Modified Body Shapes}\label{subsec:suppl_body_mod}
\begin{figure}[h]
    \centering
    \includegraphics[width=\linewidth]{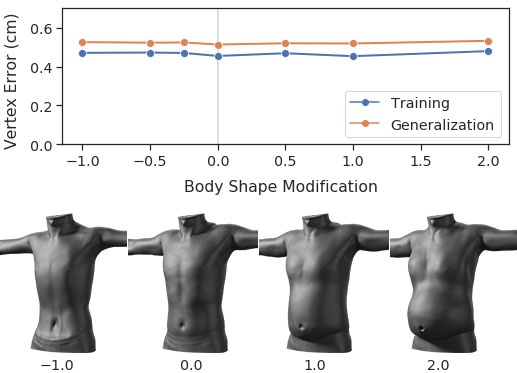}
    \caption{Training and generalization average per-vertex prediction errors (top plot) of models trained on offsets computed from different underlying body shapes (bottom row). As the body shape deviates from the true body shape (0 on the x-axis), the performance of the trained models stay roughly constant.}
    \label{fig:plot_pred_error_vs_body_deviation}
\end{figure}

In order to demonstrate the resilience of our network predictions to errors due to an incorrect assumption of the underlying body shape, we manually sculpted the scanned body and generated a number of intentionally incorrect body shapes.
With the normal body shape designated $0$ and the manually sculpted body shape designated $1$, we create a shape change parameter that ranges from $-1$ to $2$ as seen in Figure~\ref{fig:plot_pred_error_vs_body_deviation}.
The plot shows data points for 7 of our trials: the points at zero represent the original body shape, and the other 6 pairs of points represent the results obtained by training the network on the correct cloth shapes using incorrect unclothed body shape assumptions.

Also, note that the two versions of skinning with artifacts used in the paper were created on the original rigged body by manually painting weights of upper arm on the torso, and painting weights of upper arm on both the torso and the opposite arm, respectively.

\subsection{Pose Sampling}\label{subsec:suppl_poses}
\begin{figure}[h]
    \vspace{-10pt}
    \centering
    \includegraphics[width=0.75\linewidth]{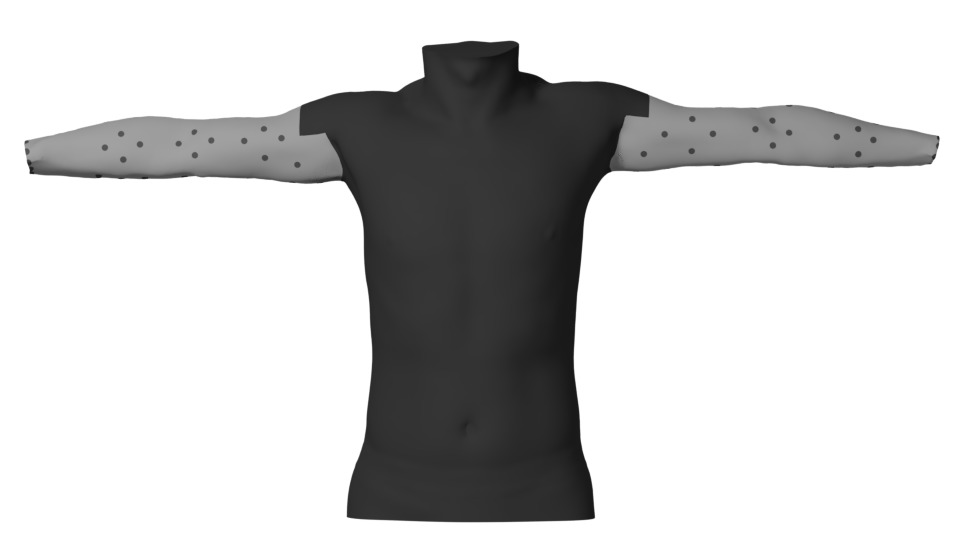}
    \caption{The body is segmented into three overlapping parts (left arm, right arm, and torso). The vertices selected for collision detection are shown as light gray dots.}
    \vspace{-10pt}
    \label{fig:body_segmentation}
\end{figure}

\begin{figure}[h]
    \centering
    \includegraphics[width=\linewidth]{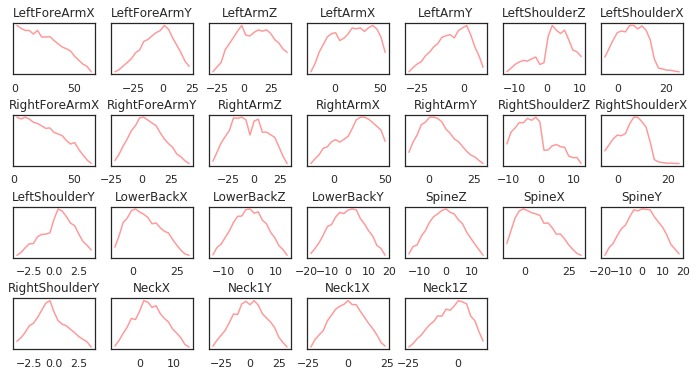}
    \caption{Plots of joint angle distributions in our dataset.}
    \label{fig:angle_distri_all}
    \vspace{-10pt}
\end{figure}

While one could sample from an empirical distribution learned from motion capture data (\eg, \cite{cmumocap}), we prefer an alternative sampling scheme in order to better cover the entire space of possible poses that can affect the T-shirt shape.
Since we only focus on the T-shirt interaction with the human body, we define the skeleton structure only for the upper body, as shown in Figure~\ref{fig:skeleton_def}. 
We fix the position and rotation for the hip (root) joint, since we are mainly interested in static poses as a first step. 
We set the joint limits according to \cite{whitmore2012nasa}, where each joint angle has both a positive limit and a negative limit for each rotation axis relative to the rest T-pose. 
For the bones on the vertical center line in Figure~\ref{fig:skeleton_def} (lower back, spine, spine1, neck, and neck1), we sample the rotation angles for each axis from a mixture of two half-normal distributions, each accounting for one direction of the rotation. 
Since we don't have such a strong prior for shoulder and arm bones, their x-axis rotation angles (azimuth) are uniformly sampled first, their z-axis rotation angles (altitude) are then uniformly sampled in the transformed space of the sine function, and finally their y-axis rotation angles are also uniformly sampled.
The rotations are applied in the order of x, z, and y.
Finally, a simple pruning procedure is applied to remove poses with severe nonphysical self-penetrations. 
This is accomplished by selecting 106 vertices from both arm parts as shown in Figure~\ref{fig:body_segmentation} and testing if any of these vertices is inside the torso.
The distributions of the sampled joint angles are shown in Figure~\ref{fig:angle_distri_all}.

\subsection{Mesh Generation}\label{subsec:suppl_meshgen}
\begin{figure}[h]
    \vspace{0pt}
    \centering
    \includegraphics[width=0.27\linewidth]{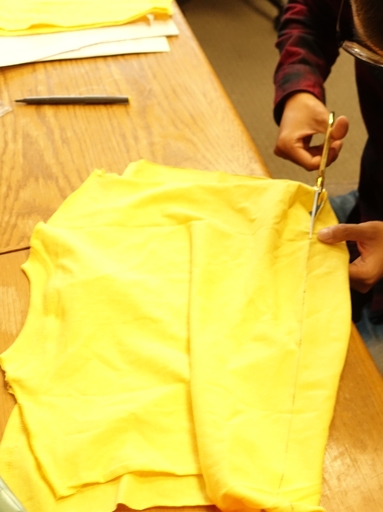}
    \includegraphics[width=0.27\linewidth]{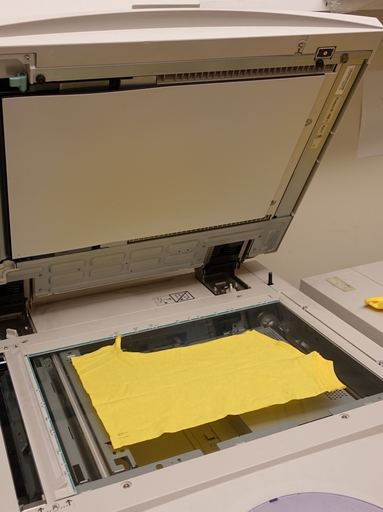}
    \includegraphics[width=0.4\linewidth,trim={0.5cm 1cm 0.5cm 1cm}, clip]{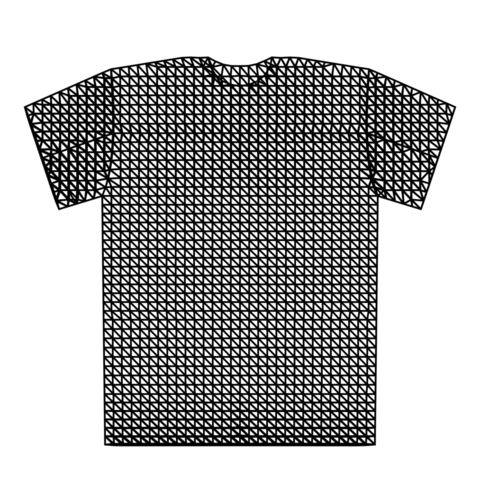}
    \caption{Illustration of the garment mesh generation process. Left: a T-shirt is cut into pieces. Middle: the pieces are scanned in. Right: the digitally stitched T-shirt mesh.}
    \vspace{0pt}
    \label{fig:mesh_gen}
\end{figure}

We carefully construct the rest state of our simulation mesh to be faithful to real-world garments by employing a reverse engineering approach where we cut up a garment along its seam lines, scan in the pieces, and then digitally stitch them back together, as shown in Figure~\ref{fig:mesh_gen}.
The T-shirt triangle mesh is 67 cm long and contains 3K vertices.
Although we try to cut the clothing into pieces such that each piece is as flat as possible to approximate flat design patterns, this may not always be achievable and the flattened versions thus obtained would not be in their intrinsic rest states leading to errors in the simulation input and distortions in the vertex UV map. 
However, such issues would be largely alleviated if one could obtain the original flat patterns from fashion designers.

\subsection{Skinning the T-shirt}\label{subsec:suppl_skinning_tshirt}
To shrink wrap the T-shirt onto the body, we first define the cage structure on both the body and the T-shirt as shown in Figure~\ref{fig:cloth_body_curves}, and then compute displacements on the T-shirt cage vertices that would morph them to the body cage; these displacement values are used as boundary conditions to solve a set of Poisson equations (see \eg \cite{Ali-Hamadi:2013,Cong:2015}) for displacements on T-shirt interior vertices.
A level set is built for the body for collision detection \cite{Bridson:2003}, and any T-shirt vertices that are inside the body are detected and pushed out to their closest points on the body surface. 

\begin{figure}
  \centering
  \includegraphics[height=0.15\textwidth]{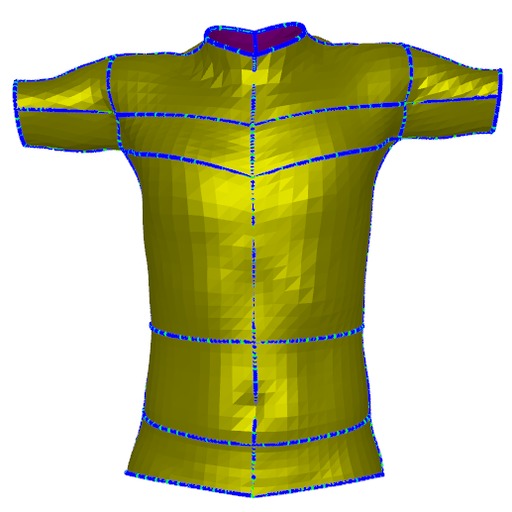}
  \includegraphics[height=0.15\textwidth]{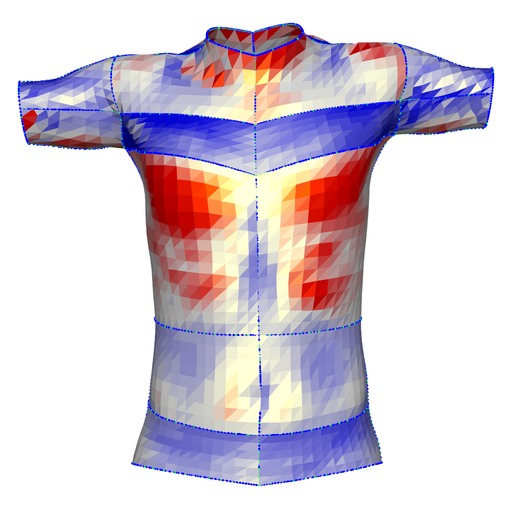}
  \includegraphics[height=0.15\textwidth]{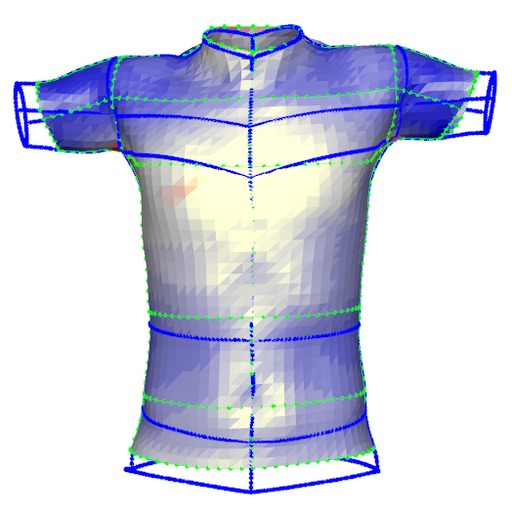}
  \caption{Shrink wrapping a T-shirt mesh onto a body in the rest pose. Left: shrink wrapped T-shirt mesh obtained by solving a Poisson equation that uses a guide ``cage'' (in blue) as a boundary condition to morph the T-shirt mesh onto the body. Middle: this initial version has area distortion, where red indicates stretching and blue indicates compression. Right: after simulation, the distortion has been reduced and more uniformly spread out so that the cloth pixels can be embedded at better locations. Note that since the T-shirt is constrained to be on the body surface, distortion is not fully eliminated.}
  \vspace{0pt}
  \label{fig:skinning_tshirt}
\end{figure}

Since the morphed T-shirt mesh can exhibit rather large and non-uniform distortion, we run a simulation using a mass-spring system to reduce distortion and achieve a better set of barycentric embedding weights for the T-shirt vertices, see Figure~\ref{fig:skinning_tshirt}.
This is done in an iterative manner. 
At each step, while constraining the T-shirt mesh to stay on the body surface, for each vertex $v$ we compute the average ratio $\bar{\alpha}_v = (1/\deg(v))\sum_{e\in E(v)}(l_e/\bar{l}_e)$ of the current edge length $l_e$ to the rest edge length $\bar{l}_e$ over its incident edges $E(v)$. 
Then for each edge $e$ with endpoints $a$ and $b$, its target edge length is set to $(1/2)(\alpha_a+\alpha_b)\bar{l}_e$.
This process essentially tries to equalize the amount of distortion for the edges incident to the same vertex, and is repeated until convergence.

\subsection{Simulation}\label{subsec:suppl_sim}
We simulate the T-shirt mesh on each sampled pose using a physical simulator~\cite{physbam} with gravity, elastic and damping forces, and collision, contact, and friction forces until static equilibrium is reached. 
To make our simulation robust to skinning artifacts that lead to cloth self-interpenetrations especially in the armpit regions, we decompose the body into three parts: the left arm, the right arm, and the torso (see Figure~\ref{fig:body_segmentation}), and associate each cloth mesh vertex with one body part as its primary collision body. 

\begin{figure}[h]
    \centering
    \includegraphics[width=0.9\linewidth,trim={1.5in 1in 0in 1in}, clip]{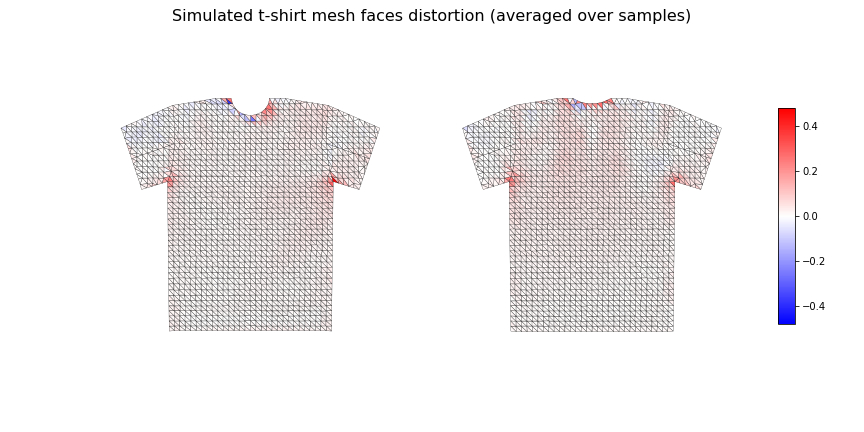}
    \vspace{0pt}
    \includegraphics[width=0.9\linewidth,trim={1.5in 1in 0in 1in}, clip]{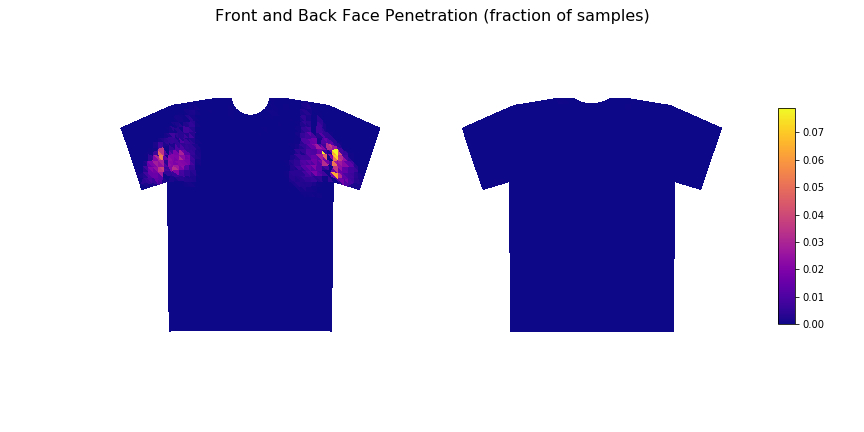}
    \caption{Mesh statistics of the simulated T-shirts. Top: front and back side average face area distortion, measured as the ratio of the current area to the rest area minus one. Bottom: front and back side per-face self-interpenetrations, measured as fraction of samples with self-interpenetrations in the dataset.}
    \vspace{0pt}
    \label{fig:mesh_dist_pene}
\end{figure}

\begin{figure}[h]
    \centering
    \includegraphics[width=\linewidth]{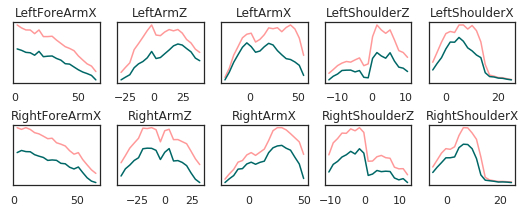}
    \caption{Visualization of selected joint angle histograms from the dataset. Red and blue lines represent the original and the filtered dataset respectively.}
    \label{fig:angle_distri_v6}
\end{figure}

After running the simulations, we further run analysis on the resulting cloth meshes and remove any shape that exhibits large distortion to reduce noise in the function to be learned. 
Specifically, if the area of any triangle in a sample compresses by more than 75\% or expands by more than 100\%, then we discard that sample. 
Figure~\ref{fig:mesh_dist_pene} shows that the amount of face area distortion is moderate (top), and the amount of self-interpenetrations is very small in the dataset (bottom).
Figure~\ref{fig:angle_distri_v6} shows that the cleaned dataset contains a similar distribution of poses as the original one.
In line with Appendix~\ref{sec:suppl_image_edit} and Figure~\ref{fig:image_edit}, one could also use image analysis on the cloth images in order to identify and prune samples that are deemed undesirable. 

\begin{figure*}[t]
 \centering
  \includegraphics[width=\linewidth]{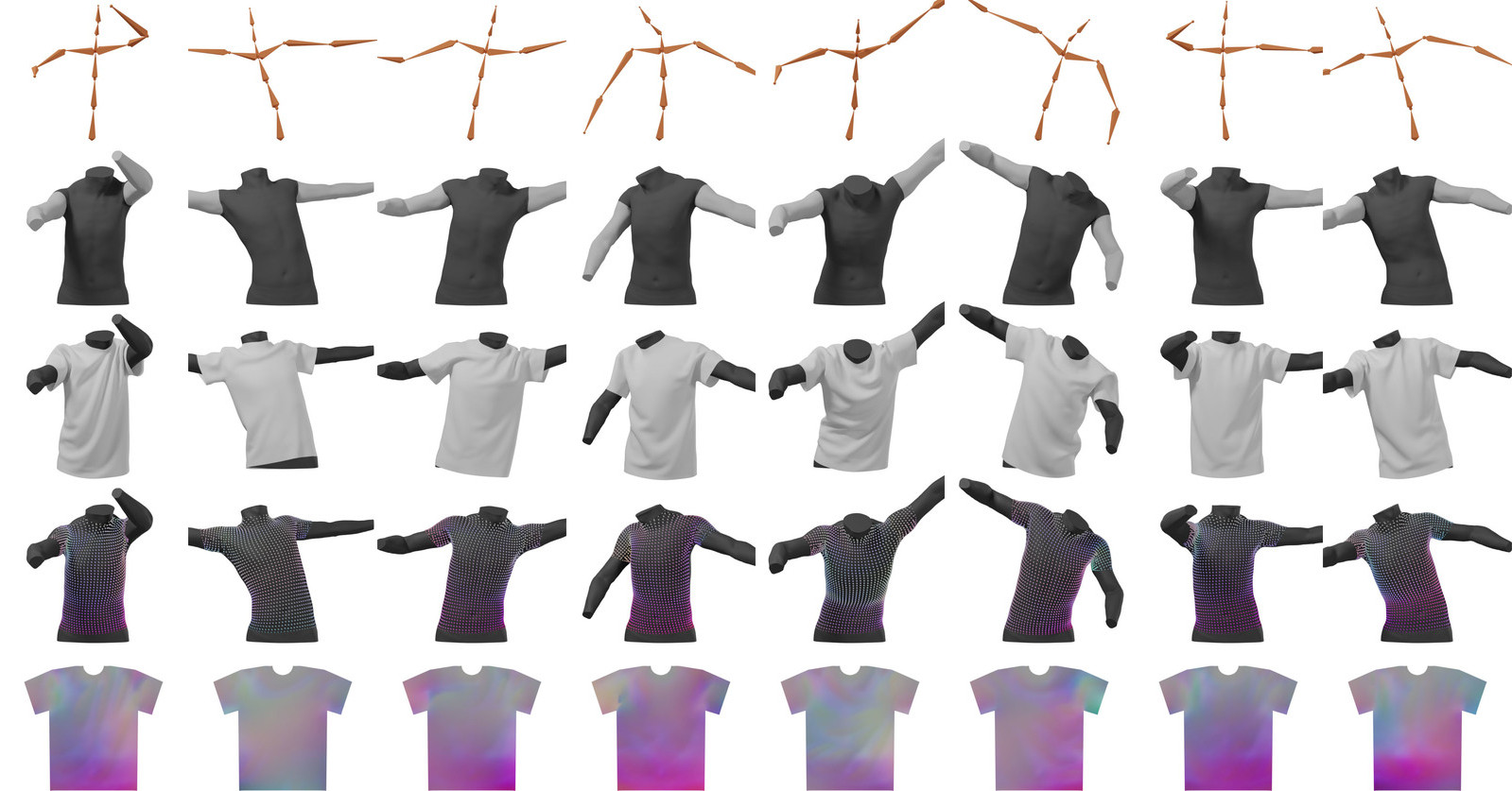}
  \caption{Random samples from our generated dataset. First row: skeletal poses. Second row: three overlapping collision bodies. Third row: simulated T-shirts. Fourth row: cloth pixels.  Fifth row: cloth images (front side).}
  \label{fig:sample_poses_data}
\end{figure*}

This leads to a total of $20{,}011$ samples that we use to train and evaluate our models, see Figure~\ref{fig:sample_poses_data} for some examples.
We create a separate UV map for the front side and the back side of the T-shirt.

\subsection{Patches}\label{subsec:suppl_patches}
There are 28 patches on the front and the back side of the T-shirt (14 each).
Whereas we train on $256\times256$ cloth images for the whole T-shirt, for each patch we make a $160\times160$ crop from a higher resolution $512\times512$ cloth image centered at the center of its axis-aligned bounding box.
The cropped patch contains $16$ pixels outside of the patch to capture the surrounding context, and the loss is computed on this enlarged patch.  

\section{Networks}\label{sec:suppl_networks}
\subsection{Architecture}\label{subsec:suppl_arch}
For predicting cloth images of the whole T-shirt, we start with the 90 dimensional input pose parameters and first apply transpose convolution followed by ReLU activation to obtain an initial $8\times8\times384$ dimensional feature map.
Then we successively apply groups of transpose convolution (filter size $4\times4$ and stride $2$), batch normalization, and ReLU activation until we reach the output resolution of $256\times256$. Each time the spatial resolution doubles and the number of channels halves. 
Finally, a convolution layer (filter size $3\times3$ and stride $1$) brings the number of channels to $6$.
The network contains 3.79 million parameters.

We use the same network architecture for all 28 patches. 
We start with the 90 dimensional input pose parameters, and first apply a linear layer to obtain a $5\times5\times512$ dimensional feature map.
Then similar to the network for the whole T-shirt, we successively apply groups of transpose convolution (filter size $4\times4$ and stride $2$), batch normalization, and ReLU activation until we reach the target resolution of $160\times160$.
Again, a final convolution layer (filter size $3\times3$ and stride $1$) brings the number of channels to $3$.
The network contains 3.96 million parameters.

\subsection{Loss Functions}\label{subsec:suppl_loss_fn}
The base loss term for grid pixel values is
\small
\begin{equation}\label{eq:base_pix_loss}
\mathcal{L}_{\mathrm{grid\_pix}}(\I^{pd},\I^{gt})=\frac{\sum_{i,j} W(i,j)||\I^{pd}(i,j)-\I^{gt}(i,j)||}{\sum_{i,j} W(i,j)},
\end{equation}
\normalsize
where $\I^{gt}$ denotes ground truth grid pixel values, $\I^{pd}$ denotes predicted grid pixel values, $W$ denotes the Boolean padded mask of the UV map, and $i,j$ are indices into the image width and height dimensions.

The additional loss term for the normal vectors is
\begin{equation}\label{eq:normal_loss}
\mathcal{L}_{\mathrm{normal}}(\I^{pd},\I^{gt})=\frac{1}{N_v} \sum_{v}(1-\mathbf{n}^{pd}_v(\I^{pd})\cdot \mathbf{n}^{gt}_v),
\end{equation}
where we compute a predicted unit normal vector $\mathbf{n}^{pd}_v$ on each vertex $v$ using the predicted grid pixel values $\I^{pd}$ (by first interpolating them back to cloth pixels and adding these per-vertex offsets to their embedded locations to obtain predicted vertex positions) and use the cosine distance to the ground truth unit normal vector $\mathbf{n}^{gt}_v$ as the loss metric. $N_v$ is the number of vertices.

Table~\ref{tab:diff_networks} shows the average per-vertex prediction errors and the normal vector errors from our convolutional decoder network trained with different loss terms on our training set and test set. The weight on the loss on normal vectors is set to $0.01$.
\begin{table}[h]
    \caption{Average per-vertex position error (in cm) and unit normal vector error (cosine distance) of our convolutional decoder network trained with different loss functions. $L_1$ and $L_2$ refer to the loss function used on the Cartesian grid pixels. N refers to normal loss.}
    \label{tab:diff_networks} 
    \centering
    \begin{tabular}{l|c|c|c|c}
        \hline
        \multirow{2}{*}{Loss} & \multicolumn{2}{c|}{Training Error} & \multicolumn{2}{c}{Generalization Error} \\
        \cline{2-5}
            & Vertex & Normal & Vertex & Normal \\
        \hline
        $L_1$ & 0.33 & 0.020 & 0.44 & 0.027 \\
        \hline
        $L_2$ & 0.35 & 0.017 & 0.47 & 0.028 \\
        \hline
        $L_2$ + N & 0.37 & 0.0075 & 0.51 & 0.029 \\
        \hline
    \end{tabular}
\end{table}
\normalsize

\subsection{Fully Connected Networks}\label{subsec:suppl_fc}
\begin{figure}[h]
    \centering
    \includegraphics[width=0.95\linewidth]{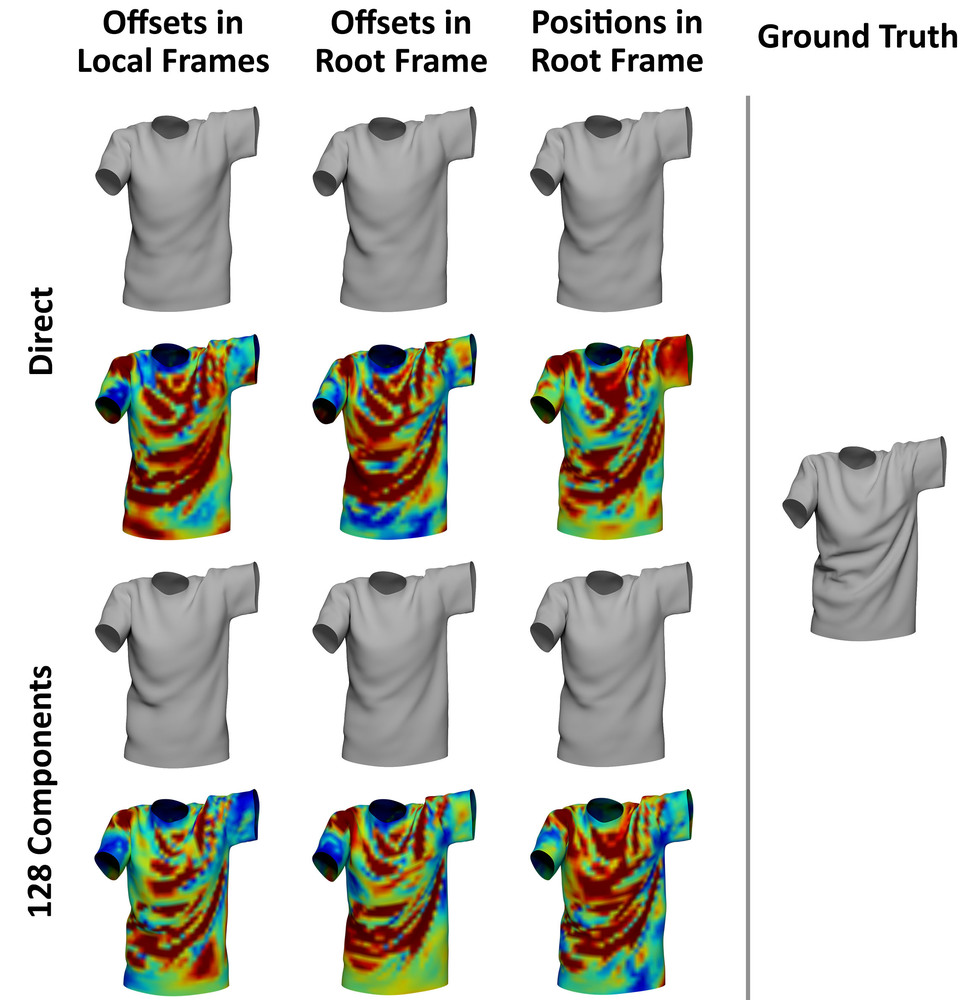}
    \caption{Comparison of fully connected network predictions and errors from models trained on different functions defined on our cloth pixels.}
    \label{fig:fc_example}
\end{figure}

We illustrate that our cloth pixel framework provides for offset functions that can be approximated via a lower dimensional PCA basis, and that a fully connected network can be trained and subsequently generalized to predict cloth shapes.
Furthermore, we compare functions of offsets represented in different spaces,
as well as functions of positions in the root joint frame.
See Table~\ref{tab:fc_pca} and Figure~\ref{fig:fc_example}.
We train a fully connected network with two hidden layers each with 256 units and ReLU activation for all the functions.
The networks trained to predict PCA coefficients indeed have better visual quality and deliver better training and generalization errors compared to the networks trained to directly predict per-vertex values.
Our experiments also show that ReLU activation leads to faster convergence and similar results compared to the Tanh activation used in \cite{Bailey:2018}.
\begin{table}[h]
    \caption{Average per-vertex position error (in cm) of the fully connected network trained with and without PCA in different spaces. ``Off\@. Loc\@.'' refers to offsets represented in local tangent-bitangent-normal frames. ``Off\@. Root\@.'' refers to offsets represented in the root joint frame. ``Pos\@. Root\@.'' refers to positions in the root frame.}
    \label{tab:fc_pca} 
    \centering
    \begin{tabular}{l|c|c}
        \hline
        Model & Training Error & Generalization Error \\
        \hline
        Off. Loc. Direct & 0.65 & 0.67 \\
        \hline
        Off. Loc. 128 PC & 0.50 & 0.55  \\
        \hline
        Off. Root. Direct & 0.69 & 0.72 \\
        \hline
        Off. Root. 128 PC & 0.53 & 0.58  \\
        \hline
        Pos. Root. Direct & 0.63 & 0.68 \\
        \hline
        Pos. Root. 128 PC & 0.58 & 0.65 \\
        \hline
    \end{tabular}
\end{table}
\normalsize

\section{Neckties}\label{sec:suppl_neckties}
Similar to the T-shirt dataset, we generate $9{,}999$ poses by randomly sampling rotation angles on 4 joints along the center line (lower back, spine, neck, and neck1), \ie, our input pose parameters are only 36 dimensional.
The dataset is divided into a training set of $7{,}999$ poses, a regularization set of $1{,}000$ poses, and a test set of $1{,}000$ poses.
In this example, we use one UV map for the entire mesh, and since the necktie has a much narrower UV map in the texture space, we modify our network architecture to predict a rectangular image with aspect ratio $1:4$ and size $64\times256$ containing 3 channels.
$L_1$ loss is used for the Cartesian grid pixels.
The weight on the normal loss is set to $0.1$.
We further add an $L_1$ loss term on the edge lengths with weight $0.1$ to ensure a smooth boundary:
\begin{equation}\label{eq:edge_loss}
\mathcal{L}_{\mathrm{edge\_len}}(\I^{pd})=\frac{1}{N_e}\sum_{e}||l_e^{pd}(\I^{pd})-l_e^{gt}||,
\end{equation}
where we compute a predicted edge length $l_e^{pd}$ for each edge $e$ using the predicted grid pixel values $\I^{pd}$ (also by first interpolating them back to cloth pixels and adding these per-vertex offsets to their embedded locations to obtain predicted vertex positions) and compare to the ground truth edge lengths $l_e^{gt}$. $N_e$ is the number of edges in the mesh.
We represent the offsets $\dx$ in the root joint frame, \ie, $(\Delta x, \Delta y, \Delta z)$, instead of the local tangent-bitangent-normal frames $(\Delta u, \Delta v, \Delta n)$.
This is more natural for the neckties, because unlike the T-shirts, they have a much larger range of displacements from the body surface while also exhibiting few high frequency wrinkles.

Since the neckties contain less high frequency variation and the output image size is smaller, a smaller network is used to learn the necktie images.
Starting from the 36 dimensional input pose parameters, we first apply a linear layer with 128 hidden units and then apply another linear layer to obtain a $8\times8\times64$ dimensional feature map.
After that, we successively apply groups of transpose convolution, batch normalization, and ReLU activation as above until we reach the target resolution of $64\times256$.
Then, a final convolution layer (filter size $3\times3$ and stride $1$) brings the number of channels to $3$.
The network contains 2.16 million parameters.

\newpage
{\small
\bibliographystyle{ieee}
\bibliography{references}
}

\end{document}